\documentclass[a4paper]{cas-sc}

\usepackage[authoryear]{natbib}
\usepackage{amsthm}
\usepackage{amsmath}
\usepackage{subfigure}
\newtheorem{definition}{Definition}[section]
\newtheorem{lemma}{Lemma}
\def\tsc#1{\csdef{#1}{\textsc{\lowercase{#1}}\xspace}}
\tsc{WGM}
\tsc{QE}
\tsc{EP}
\tsc{PMS}
\tsc{BEC}
\tsc{DE}

\begin{document}
\let\WriteBookmarks\relax
\def\floatpagepagefraction{1}
\def\textpagefraction{.001}
\shorttitle{A multi-source data power load forecasting method using attention mechanism-based parallel cnn-gru}
\shortauthors{Chao Min et~al.}

\title [mode = title]{A multi-source data power load forecasting method using attention mechanism-based parallel cnn-gru}

\author[swpu1,swpu2,swpu3]{Chao Min}[orcid=0000-0003-1245-8142]\corref{mycorrespondingauthor}\ead{minchao@swpu.edu.cn}
\author[swpu1,swpu2]{Yijia Wang}
\author[swpu4]{Bo Zhang}
\author[swpu5]{Xin Ma}
\author[swpu1,swpu2]{Junyi Cui}
\address[swpu1]{School of Science, Southwest Petroleum University, Chengdu 610500, China}
\address[swpu2]{Institute for Artificial Intelligence, Southwest Petroleum University, Chengdu 610500, China}
\address[swpu3]{State Key Laboratory of Oil and Gas Reservoir Geology and Exploitation, Chengdu 610500, China}
\address[swpu4]{Sinopec Zhongyuan Oilfield Company, Puyang 457000, China}
\address[swpu5]{School of Mathematics and Physics, Southwest University of Science and Technology, Mianyang 621000, China}

\cortext[mycorrespondingauthor]{Corresponding author at: School of Science, Southwest Petroleum University, Chengdu 610500, China.}

\begin{abstract}
Accurate power load forecasting is crucial for improving energy efficiency and ensuring power supply quality.
Considering the power load forecasting problem involves not only dynamic factors like historical load variations but also static factors such as climate conditions that remain constant over specific periods.
From the model-agnostic perspective, this paper proposes a parallel structure network to extract important information from both dynamic and static data.
Firstly, based on complexity learning theory, it is demonstrated that models integrated through parallel structures exhibit superior generalization abilities compared to individual base learners.
Additionally, the higher the independence between base learners, the stronger the generalization ability of the parallel structure model.
This suggests that the structure of machine learning models inherently contains significant information.
Building on this theoretical foundation, a parallel convolutional neural network (CNN)-gate recurrent unit (GRU) attention model (PCGA) is employed to address the power load forecasting issue, aiming to effectively integrate the influences of dynamic and static features.
The CNN module is responsible for capturing spatial characteristics from static data, while the GRU module captures long-term dependencies in dynamic time series data.
The attention layer is designed to focus on key information from the spatial-temporal features extracted by the parallel CNN-GRU.
To substantiate the advantages of the parallel structure model in extracting and integrating multi-source information, a series of experiments are conducted.
The performance of the PCGA model is compared against baseline models and serially constructed models, and a detailed error analysis on the prediction results of each model ia performed.
The results demonstrate that the PCGA model significantly improves prediction accuracy by effectively extracting the spatial and temporal characteristics of the relevant factors of power load forecasting.
\end{abstract}


\begin{highlights}
\item Demonstrating the advantages of parallel network architectures using complexity learning theory.
\item Constructing a parallel cnn-gru attention model for power load prediction.
\item Our model can effectively extract the spatiotemporal features relevant to power load forecasting.
\item Experiments show that the prediction accuracy of our model is significantly improved compared with others.
\end{highlights}


\maketitle

\section{Introduction}
Given the continuous growth in societal electricity demand, ensuring the stability and quality of power supply has become an increasingly urgent task \citep{2022Providing}.
This imposes higher requirements on the secure and efficient operation of the power grid, with accurate power load forecasting emerging as a key solution to this challenge \citep{Gao2022MidtermED}.
Moreover, accurate prediction of power load is crucial for prompt response and real-time scheduling of power systems \citep{Pansota2021SchedulingAS}.
Thus, achieving reliable operation and optimized scheduling of the power system hinges significantly on precise power load forecasting \citep{2020Towards}.

However, the task of power load forecasting is influenced by various external factors, such as human socio-economic activities, meteorological conditions, holidays, and regional disparities \citep{2022Fisher}.
Additionally, the load data exhibits nonlinearity, time-variation, and uncertainty, which can lead to a decrease in the generalization ability of machine learning models and complexities in model interpretation, posing challenges in accurately forecasting power load \citep{Tang2023ShortTermPL}.
Conventional load forecasting methods fail to effectively uncover the relationship between load data and various factors.
Therefore, employing an appropriate model to fully leverage the latent features in the load sequence is crucial for enhancing prediction accuracy.

\subsection{Related works}\label{section: Related Works}
Currently, many scholars primarily employ three kinds of approaches to investigate load forecasting: statistical analysis, physical model, and machine learning.
Statistical methods generally include time series analysis and regression analysis.
Commonly used time series methods such as ARMA \citep{Wang2016ImprovedST,Bizrah2017TheIO} and ARIMA \citep{Singhal2019ShortTermLF,Wenlong2020LoadFO}, etc., analyze historical power load data to detect patterns like periodicity, trends, and seasonality for prediction purposes.
Regression analysis establishes relationships between power load and influencing factors (e.g., temperature, time) to predict outcomes via regression models \citep{Madhukumar2022RegressionMS,Luo2017ShortTP}.
These methods are simple in principle, fast in execution, and capable of identifying data's time-varying patterns.
However, due to oversimplification, these methods struggle to handle the complex nature of load systems affected by multiple factors. They are unable to effectively address nonlinear and uncertainty issues, and may lead to under-fitting.

The physical models incorporate the electric power system's physical characteristics and load curves to establish corresponding predictive models.
A novel physics-based building stock energy model (BSEM) framework \citep{Kim2023PhysicsbasedMO} has been proposed for simulating electricity load curves in commercial sectors.
This innovation addresses significant errors in traditional modeling approaches arising from inadequate consideration of large population and heterogeneity in building systems.
Given the similarity of power demand patterns of individual buildings and the advantages of generative adversarial network (GAN), a new approach (E-GAN) \citep{Tian2022Dailypower} combining a physics-based model (EnergyPlus) with a data-driven model (GAN), is proposed to accurately forecast daily power demand for buildings at a large scale.
The physical model is based on the physical law of the power system, which can provide an explanation for changes in power load, help to understand load variation mechanisms, and generally ensure robust stability.
However, compared to statistical methods, constructing physical models involves substantial workload and lengthy computation times.
Additionally, it is difficult for physical models to accurately capture complex nonlinear relationships.

Compared to statistical-based and physical-based methods, machine learning methods are effective in dealing with nonlinear problems \citep{Zhang2018ANR}.
The traditional machine learning methods commonly used to predict power load include KNN \citep{Lv2018Shortterm}, decision tree \citep{Qi2017LoadPR}, support vector regression \citep{Nazeer2019ShortTerm}, random forest \citep{Dang2022AQR}, and their hybrid models \citep{Yamasaki2024OptimizedHE}.
Deep learning technology has been widely applied in power load forecasting due to its robust nonlinear mapping capability and self-adaptability \citep{Ibrahim2022MachineLF}.
For instance, ANN \citep{Belhaiza2024NeuralNetwork}, RNN \citep{Smyl2022ESdRNNWD}, LSTM \citep{Song2024AML}, CNN \citep{Sadaei2019ShorttermLF}, and other representative deep neural network algorithms
In addition, there are some hybrid deep learning frameworks \citep{Chen2024DayaheadLF,Kim2019PredictingRE,Sajjad2020ANC} are employed to tackle the nonlinear distribution challenges with high volatility, uncertainty, and random characteristics \citep{Alghamdi2023AnIM}.
However, machine learning methods are sometimes referred to as "black boxes" because the process of model learning from data and making predictions is not easily interpretable, thus leading to the construction of many learning frameworks with inherent opacity \citep{Arrieta2019ExplainableAI}.

To improve the interpretability of the model construction process, the approach is made from the perspective of aligning models with data, with different machine learning model modules employed to handle diverse types and sources of data.
These modules are integrated through a parallel structure.
Analysis of model complexity has shown that this parallel structure effectively improves the model's generalization ability.
For the task of power load forecasting, a CNN-GRU attention parallel network framework based on multi-source data fusion is propoesd.
This framework simultaneously captures spatial information and temporal correlations within the data, adapting more flexibly to variations of data in different scenarios, and making better use of data diversity and heterogeneity.
This approach introduces a more comprehensive and accurate modeling approach for power load forecasting.

\subsection{Contribution}\label{section: Contribution}
The main contributions of this study are listed as follows:
\begin{enumerate}[1)]
\item The theoretical basis for constructing parallel form models is presented, and complexity learning theory is employed to demonstrate that the generalization ability of the model integrated through parallel structure is significantly enhanced compared to single machine learning modules (i.e., individual base learners).
    It is further demonstrated that greater independence between the base learners leads to improved generalization ability of the integrated model.
    Building on these theoretical foundations, a parallel CNN-GRU attention model for power load forecasting is proposed.
    This model integrates CNN and GRU using a parallel architecture to effectively capture the spatiotemporal characteristics of multivariate time series data, thereby improving the accuracy of power load forecasting.
\item The factors influencing power load forecasting are comprehensively constructed and analyzed, with input features screened through data-driven methods.
    To identify the relevant input factors, various correlation coefficients and multivariate correlation measures based on Copula entropy are utilized.
    Additionally, Granger causality is applied to determine the causal relationships between each input factor and power load, clarifying the model inputs and thereby improving model accuracy.
\item In order to verify the effectiveness of the proposed model, the model with superior prediction performance is identified from a comparison of five shallow baseline models (BP, CNN, LSTM, BILSTM, GRU).
    Subsequently, six new models (SCL, PCL, SCG, PCG, SCGA, PCGA) are constructed using serial and parallel methods.
    Performance comparisons are conducted using MAPE, RMSE, MAE and R2.
    Additionally, an error analysis of the prediction results is conducted.
    It is demonstrated that the prediction accuracy of the PCGA model has been significantly improved, and the spatial and temporal characteristics related to power load forecasting are effectively extracted.
\end{enumerate}

The rest of this paper is organized as follows.
Section \ref{section: Theoretical Analysis of Generalization ability in Parallel Fusion Frameworks} presents the theoretical foundation for constructing parallel-form models.
Section \ref{section: The proposed parallel cnn-gru attention model} elaborates on the modeling process.
Section \ref{section: Preparation of case study} details the preparation for the case study, including the experimental metrics, data preparation, and an overview of the base models.
Section \ref{section: Experimental result} outlines the experimental details and results.
Finally, Section \ref{section: Conclusion} summarizes the main conclusions of the study.

\section{Theoretical analysis of generalization ability in parallel fusion frameworks}\label{section: Theoretical Analysis of Generalization ability in Parallel Fusion Frameworks}
This section will apply complexity learning theory to demonstrate the enhancement of generalization ability in parallel form models, as well as the facilitating effect of combining independent base learners on generalization performance.
Through rigorous theoretical derivation and analysis, it is argued that the advantages of parallel form models over the single model in generalization ability are highlighted, which helps to deepen the understanding of the performance improvement mechanism of the parallel form model.

\subsection{Complexity learning theories}\label{section: Complexity Learning Theories}
Firstly, we introduce the Rademacher complexity in complexity learning theory and the generalized error boundary theorem based on Rademacher complexity, which provides the necessary basis for further theoretical derivation and proof.

\vspace{7pt}
a) Rademacher complexity

Rademacher complexity is a way to characterize the complexity of the hypothesis space, which is tighter than the results of the learnability analysis based on VC dimension, and considers the distribution of data to a certain extent.
The definition of Rademacher complexity is given below.
\begin{definition}[\citealp{2003Machine}]
Consider the real-valued function space $\mathcal{H}:\mathcal{D}\to\mathbb{R}$, let $D=(x_i,y_i)_{i=1}^n$, where $(x_i,y_i)\in\mathcal{D}$.
Suppose that $\sigma=\{\sigma_1,\sigma_2,...,\sigma_n\}$ is a sample containing n independent and identically distributed Rademacher random variables, where $\sigma_{i}$ takes the values -1 and +1 with equal probability of 0.5, referred to as Rademacher random variables.
For the regression problem, let the loss function be $\ell(f(x),y)$, and the empirical Rademacher complexity $\widehat{\mathcal{R}}_n(\mathcal{H})$ of $D$ is defined as:
\begin{equation}
\widehat{\mathcal{R}}_n(\mathcal{H})=\mathbf{E}_\sigma\bigg[\sup_{f\in \mathcal{H}}\frac{1}{n}\sum_{i=1}^n\sigma_i\ell\big(f(x_i),y_i\big)\bigg],
\label{equ7}
\end{equation}
then the Rademacher complexity of the function space $\mathcal{H}$ is $\mathcal{R}_n(\mathcal{H})=\mathbf{E}\widehat{\mathcal{R}}_n(\mathcal{H})$.
\end{definition}

\vspace{7pt}
b) The generalization error bounds theorem based on Rademacher complexity

Firstly, let's introduce some notations.
The input space $\mathcal{X}$, the action space $\mathcal{A}$, the output space $\mathcal{Y}$, and a class of functions $F$ are given.
There is a loss function $\ell:\mathcal{A}\times\mathcal{Y}\to[0,1]$, for any $y\in\mathcal{Y}$ and $a\in\mathcal{A}$, $\ell(y,a)$ reflects the cost of taking specific action $a\in\mathcal{A}$ when the outcome is $y\in\mathcal{Y}$.
The control cost function $\phi:\mathcal{Y}\times\mathcal{A}\to\mathbb{R}$ is used to control the loss function $\ell$, for any $y\in\mathcal{Y}$ and $a\in\mathcal{A}$, $\phi(y,a)\geq\ell(y,a)$.
If $\phi$ is a function defined on the range of functions in $F$, let $\phi\circ F=\{\phi\circ f\mid f\in F\}$.
Given an independent sample $(X_i,Y_i)_{i=1}^n$, the distribution is $(X,Y)$.
The objective of learning is to select a function $f\in F$ that maps from $\mathcal{X}$ to $\mathcal{A}$ to minimize the expected loss $\mathbf{E}\ell(Y,f(X))$.
A generalization error bound theorem based on Rademacher complexity is given below.
\begin{lemma}[\citealp{Bartlett2003RademacherAG}]
\label{lemma1}
Consider the loss function $\ell:\mathcal{A}\times\mathcal{Y}\to[0,1]$ and the control cost function $\phi:\mathcal{Y}\times\mathcal{A}\to[0,1]$.
Let $F$ be a class of function mappings from $\mathcal{X}$ to $\mathcal{A}$, and $(X_i,Y_i)_{i=1}^n$ is independently selected according to the probability measure $P$.
For any integer $n$ and any $0<\delta<1$ (where $\delta$ is used to control the confidence level for estimating the complexity of the hypothesis space), the probability on the sample with length $n$ is at least $1-\delta$, and every $f$ in $F$ satisfies:
\begin{equation}
\mathbf{E}\ell(Y,f(X))\leq\hat{\mathbf{E}}_n\phi(Y,f(X))+\mathcal{R}_n(\tilde{\phi}\circ F)+\sqrt{\frac{8\ln(2/\delta)}n},
\label{equ8}
\end{equation}
where $\tilde{\phi}\circ F=\{(x,y)\mapsto\phi(y,f(x))-\phi(y,0):f\in F\}$.
\end{lemma}
\begin{lemma}[\citealp{Xu2017InformationtheoreticAO}]
\label{lemma2}
Given two random variables $X$ and $Y$ with joint distribution $P_{X,Y}$, if $f(X,Y)$ satisfies $\sigma\mathrm{-subgaussian}$ under the marginal probability density function $P_{\overline{X},\overline{Y}}=P_{X}\otimes P_{Y}$, then
\begin{equation}
    \left|\mathbf{E}_{_{(X,Y)}}[f(X,Y)]\mathbf{-}\mathbf{E}_{_{X\otimes Y}}[f(X,Y)]\right|\leq\sqrt{2\sigma^{2}I(X;Y)}.
\label{equ9}
\end{equation}

Note: If a random variable $U$ satisfies $\log\mathbf{E}[e^{\lambda(U-\mathbf{E}[U])}]\leq\frac{\lambda^2\sigma^2}2$ for $\lambda\in\mathbb{R}$, then $U$ is termed as $\sigma\mathrm{-subgaussian}$.
\end{lemma}

\subsection{Enhancement of the generalization ability in parallel structure models}\label{section: Enhancement of the Generalization Ability in Parallel Structure Models}
This paper evaluates the generalization ability of the parallel model by presenting the generalization error bounds of the model.
In machine learning models, the loss function often exhibits high complexity, involving multiple variables and intricate nonlinear interactions.
Due to this complexity, the optimal solution is typically not unique, and the training process may encounter multiple local optimal solutions.
The relationships between these local optimal solutions depend on the form of the loss function and the characteristics of the optimization algorithm.
Despite some local optima potentially deviating from the global optimum, in practical applications, certain suboptimal local optima can still provide satisfactory predictive results.
This is due to the outstanding fitting ability of machine learning models, enabling them to approximate the global optimum using locally optimal parameter settings.

Firstly, necessary symbols are introduced for theoretical proof.
Consider a learning task on dataset $(x,y)\in\mathcal{X}\times\mathcal{Y}$, where input data $x=\{x^{(1)},\cdots,x^{(M)}\}$ originates from $M$ distinct data sources, with $x^{(i)}$ potentially in vector or matrix form, and $y$ representing output data.
The training dataset is defined as $D_{\mathrm{train}}=\{x_{i},y_{i}\}_{i=1}^{N}$.
${\mathcal{X}}$, ${\mathcal{Y}}$, and ${\mathcal{Z}}$ denote the input space, target space, and potential space, respectively.
$h{:}\mathcal{X}\mapsto\mathcal{Z} , h=(h_{1},...,h_{M})$ is a multi-source data fusion mapping from input space to potential space.
$g\colon\mathcal{Z}\mapsto\mathcal{Y}$ is a task mapping.
The objective is to learn a reliable multi-source data fusion model $f=g\circ h(x)$ that performs well on an unknown test dataset $D_{\mathrm{test}}$.
$D_{\mathrm{train}}$ and $D_{\mathrm{test}}$ are both from the joint distribution ${\mathcal{D}}$ on $\mathcal{X}\times\mathcal{Y}$.
Here, $f=g\circ h(x)$ represents the composite function of $h$ and $g$.

In the case of containing $M$ data sources, $f^{m}$ is defined as the base learner on data source $x^{(m)}( m=1,2,\cdots,M)$.
Let $\mathcal{W}$ denote the hypothesis space of model parameters for multi-source data fusion, where $W\in\mathcal{W},W=(w^{1},\cdots,w^{M})$ represents specific network parameters given as a function of input sample $x$.
The final output value is computed through $f(x)=\sum_{m=1}^{M}w^{m}\cdot f^{m}(x^{(m)})$.

Let $(x,y)\sim\mathcal{D}$ denote the multi-source data sample, and the loss function of the model is defined as:
\begin{equation}
    \ell(f(x),y)=\ell(\sum_{m=1}^{M}w^{m}\cdot f^{m}(x^{(m)}),y).
\label{equ10}
\end{equation}

The generalization error of the multi-source data fusion model is defined as:
\begin{equation}
    GError(f)=\mathbf{E}_{(x,y)\sim\mathcal{D}}[\ell(f(x),y)],
\label{equ11}
\end{equation}
where $\ell(f^{m}(x^{(m)}),y)$ represents the loss of the base learner, simplified to $l^{m}$, and omitted in the subsequent analysis.

Consider $D_{train}=\{x_{i},y_{i}\}_{i=1}^{N}$ as the training dataset comprising N samples, with $\hat{E}(f^m)$ representing the empirical error of the base learner $f^{m}$ on $D_{\mathrm{train}}$.
Given a hypothesis space containing all possible regression functions, $f$ represents any hypothesis in the hypothesis space $\mathcal{H}$.

Due to the combined effect of local convexity, convex combination, and fixed parameters of machine learning loss functions, under certain conditions, the optimization problem of machine learning models can be simplified to a convex optimization problem.
Therefore, to facilitate analysis, the loss function is approximated as convex.
Thus:
\begin{equation}
    \ell(f(x),y)=\ell(\sum_{m=1}^{M}w^{m}f^{m}(x^{(m)}),y)\leq\sum_{m=1}^{M}w^{m}\ell(f^{m}(x^{(m)}),y).
\label{equ12}
\end{equation}

Then take the expected value of both sides of the above equation:
\begin{equation}
    \mathbf{E}_{(x,y)\sim\mathcal{D}}\ell(f(x),y)\leq\mathbf{E}_{(x,y)\sim\mathcal{D}}\sum_{m=1}^{M}w^{m}\ell(f^{m}(x^{(m)}),y).
\label{equ13}
\end{equation}

Since the expectation is a linear operator, the expectation of a product equals the product of the expectation and the covariance.
Therefore, we can further decompose the right side of the equation as:
\begin{equation}
\begin{aligned}
\mathbf{E}_{(x,y)\sim\mathcal{D}}\ell(f,y)& \leq\sum_{m=1}^{M}\mathbf{E}_{(x,y)\sim\mathcal{D}}[w^{m}\ell(f^{m},y)] \\
&=\sum_{m=1}^{M}[\mathbf{E}_{(x,y)\sim\mathcal{D}}(w^{m})\mathbf{E}_{(x,y)\sim\mathcal{D}}(\ell(f^{m},y))+Cov(w^{m},\ell(f^{m},y))].
\end{aligned}
\label{equ14}
\end{equation}

According to Lemma \ref{lemma1}, for any hypothesis $f^{m}$ (i.e., $\mathcal{H}{:}\mathcal{X}\to\{-1,1\},f\in\mathcal{H}$) in $\mathcal{H}$, the probability of at least $1-\delta(1>\delta>0)$ satisfies:
\begin{equation}
    \mathbf{E}_{(x,y)\sim\mathcal{D}}(f^m)\leq\hat{E}(f^m)+\mathcal{R}_m(f^m)+\sqrt{\frac{ln(1/\delta)}{2N}},
\label{equ15}
\end{equation}
this formula represents the generalization error bound of an individual base learner, and $\mathcal{R}_{m}(f^{m})$ denotes the Rademacher complexity.

From Eq. \eqref{equ14} and \eqref{equ15}, we deduce that the following inequality holds with probability at least:
\begin{equation}
\begin{aligned}
    GError(f)&
    \leq\underbrace{\sum_{m=1}^M\mathbf{E}(w^m)\hat{E}(f^m)}_{\text{Term-L(average empirical loss)}}+\underbrace{\sum_{m=1}^M\mathbf{E}(w^m)\mathcal{R}_m(f^m)}_{\text{Term-C(average complexity)}}
    +\underbrace{\sum_{m=1}^MCov(w^m,l^m)}_{\text{Term-Cov(cowariance)}}+M\sqrt{\frac{ln(1/\delta)}{2N}},
\end{aligned}
\label{equ16}
\end{equation}
this formula is the generalization error boundary of the multi-source fusion model.

By analyzing and comparing the generalization error bounds of an individual base learner (Eq. \eqref{equ14}) and the fused model (Eq. \eqref{equ15}), it is demonstrated that the fused model exhibits improved generalization abilities compared to an individual base learner.
The generalization error bound of the fused model encompasses four components: weighted average empirical error (Term-L), Rademacher complexity (Term-C), covariance (Term-Cov), and confidence interval error.
To compare the two error bounds, we assume that each base learner $f^m$ within the fusion model performs similarly when used individually as it does within the fused model, that is, indicating similar empirical errors and complexity terms.
The first two terms of the fusion model's error bound incorporate the weight $\mathbf{E}(w^{m})$ indicating the contribution of individual base learner within the fusion model.
By judiciously assigning weights, the fusion model can emphasize well-performing base learners while suppressing poor-performing ones, aiming to reducing the overall generalization error.
The covariance term signifies conflicts or inconsistencies among different base learners.
By employing an appropriate fusion strategy, these conflicts can be minimized, thereby further reducing generalization error.
The confidence interval error term introduces additional uncertainty due to the utilization of multiple base learners.
However, its impact diminishes gradually as the sample size $N$ increases sufficiently \citep{Zhang2023ProvableDF}.

Based on the above analysis, through reasonable weight distribution and fusion strategy, the fusion model could potentially achieve a lower generalization error bound compared to individual base learners.
However, this does not imply that the fusion model is always superior to individual learners, as actual effectiveness also hinges on factors such as the quality of base learners, choice of fusion strategy, and characteristics of the dataset.

Therefore, to demonstrate that the generalization error bound of a parallel multi-source data fusion model is lower than that of a single base learner, further analysis of the specific implementation of fusion strategy and interactions among base learners is required.

\subsection{Enhanced generalization ability through ensemble of independent base learners}\label{section: Enhanced Generalization ability Through Ensemble of Independent Base Learners}
To analyze the impact of interactions between base learners on model generalization ability, a new generalization error boundary was employed.
This approach aims to demonstrate that higher independence among base learners enhances the model's generalization ability.
Inspired by ensemble learning concepts, combining multiple weak learners forms a stronger learner with improved generalization ability.
When base learners are independent, they can capture more complementary features, thereby boosting overall model generalization performance.

Firstly, let $X$, $Y$, and $Z$ denote the input space, target space, and potential space respectively, where the potential space represents the underlying features extracted by base learners, typically denoting a low-dimensional vector space.
Consider the dataset $(x,y)\in(X,Y)$, where $x=\{x^{(1)},\cdots,x^{(M)}\}$ consists of $M$ distinct data sources and $y\in Y$ represents data labels.
Additionally, let $D=(X,Y)$ denote the sample space, and dataset $S$ denote a tuple $S=(D_{1},...,D_{N})$ of size $N$, where independently and identically distributed variable $D_{i}\in D$ is sampled from an unknown distribution $P_{s}$.
Let $\mathcal{W}$ represent the hypothesis space of multi-source data fusion model parameters.
Given $W\in\mathcal{W}$, $W=(w^{1},\cdots,w^{M})$ represents the specific network parameters, and $W$ is the function of the input sample $x$.
Define $h\colon\mathcal{X}\mapsto\mathcal{Z}$, $h=(h_{1},...,h_{M})$ is the multi-source data fusion mapping from input space to potential space.
Let $g:\mathcal{Z}\mapsto\mathcal{Y}$ denote the task mapping.
Let $\hat{h}_{i}(x)=g\circ h_{i}(x)$, then $\hat{h}=(\hat{h}_1,...,\hat{h}_M)$.
Let $f(D)=(\hat{h}(X),Y)$, $f(S)=(f(D_{1}),...,f(D_{N}))$.

Assuming $W=(w^{1},\cdots,w^{M})\in\mathcal{W}$, the empirical risk on dataset $S$ can be defined as:
\begin{equation}
    L_{f(S)}(W)=\frac{1}{N}\sum_{i=1}^{N}\ell(f(D_{i}),W).
\label{equ17}
\end{equation}

The overall risk of $W$ on distribution $P_{s}$ is:
\begin{equation}
\begin{aligned}
L_{\overline{f(S)}}(W) =\mathbf{E}_{S}[\frac{1}{N}\sum_{i=1}^{N}\ell(f(D_{i}),W)]=\mathbf{E}_{f(S)}[\frac{1}{N}\sum_{i=1}^{N}\ell(F_{i},W)],
\end{aligned}
\label{equ18}
\end{equation}
it reflects the model's generalization ability under the true distribution.
Here, $F_{i}=f(D_{i})(1\leq i\leq N)$ is an independently and identically distributed random variable, with its continuity and discreteness depend on the mapping function $h_{i}$ and its corresponding data source $x^{(i)}$.

For a learning algorithm characterized by $P_{W|S}$, its generalization error is the difference $L_{\overline{f(S)}}(W) -L_{f(S)}(W)$ between Eq. \eqref{equ17} and \eqref{equ18}, and its expected value is represented as:
\begin{equation}
\begin{aligned}
g(P_{S},P_{W|S})&=\left|\mathbf{E}_{(S,W)}[L_{\overline{f(S)}}(W)-L_{f(S)}(W)]\right|\\&=\left|\mathbf{E}_{(f(S),W)}[L_{f(S)}(W)-\mathbf{E}_{f(S)}[L_{f(S)}(W)]]\right|\\&=\left|\mathbf{E}_{(f(S),W)}[L_{f(S)}(W)]-\mathbf{E}_{f(S)\otimes W}[L_{f(S)}(W)]\right|,
\end{aligned}
\label{equ19}
\end{equation}
where $\otimes$ denotes the joint distribution, and $\mathbf{E}_{f(S)\otimes W}$ denotes the expectation of the product of the marginal probability density functions $f(S)$ and $W$.
The absolute value is applied because the model's performance on different datasets could be positive or negative; using the absolute value eliminates this polarity difference, making it easier to understand the overall performance variation of the learning.

If the loss functions in Eq. \eqref{equ17} and \eqref{equ18} are constrained within the range $[a,b]$, then $L_{f(S)}(W)$ in Eq. \eqref{equ19} is $\sigma/\sqrt{N}-subgaussian$ for $W$, where $\sigma=(b-a)/2$.
This step utilizes the Chernoff method \citep{Chernoff1952A} to estimate the moment generating function of $L_{f(S)}(W)$.
According to Lemma \ref{lemma2}, the upper bound of the generalization error is expressed as the mutual information between $f(S)$ and $W$:
\begin{equation}
    g(P_{S},P_{W|S})\leq\frac{1}{N}\sqrt{2\sigma^{2}}\sqrt{NI(f(S);W)}.
\label{equ20}
\end{equation}

Let $S_{x}=\{X_{1},...,X_{N}\}$, $S_{y}=\{Y_{1},...,Y_{N}\}$.
To quantify $g(P_{S},P_{W|S})$, decompose $I(f(S);W)$:
\begin{equation}
\begin{aligned}
I(f(S);W)& =I(\hat{h}(S_{x}),S_{y};W) \\
&=I(W;\hat{h}(S_{x}))+I(S_{y},W | \hat{h}(S_{x})) \\
&=-I(S_{y};\hat{h}_{1}(S_{x}),\hat{h}_{2}(S_{x}),...,\hat{h}_{M}(S_{x}))+H(S_{y})+H(W)-H(S_{y},W\mid\hat{h}(S_{x})) \\
&=I(\hat{h}_{1}(S_{x}),\hat{h}_{2}(S_{x}),...,\hat{h}_{M}(S_{x}))-I(\hat{h}_{1}(S_{x}),\hat{h}_{2}(S_{x}),...,\hat{h}_{M}(S_{x})|S_{y})-\sum_{x=1}^mI(S_y,\hat{h}_i(S_x))\\
&\quad+H(S_y)+H(W)-H(S_y,W|\hat{h}(S_x)).
\end{aligned}
\label{equ21}
\end{equation}

\begin{equation}
    \begin{aligned}
I(\hat{h}_{1}(S_{x}),\hat{h}_{2}(S_{x}),...,\hat{h}_{M}(S_{x})|S_{y})
&=\mathbf{E}_{(\hat{h}(S_{x}),S_{y})}\log\frac{P(\hat{h}_{1}(S_{x}),...,\hat{h}_{M}(S_{x})|S_{y})}{P(\hat{h}_{1}(S_{x})|S_{y})P(\hat{h}_{2}(S_{x})|S_{y}),...,P(\hat{h}_{M}(S_{x})|S_{y})} \\
&=\mathbf{E}_{(\hat{h}(S_{x}),S_{y})}\log\frac{\prod_{i=1}^{N}P(\hat{h}(X_{i}) | Y_{i})}{\prod_{i=1}^{N}P(\hat{h}_{1}(X_{i}) | Y_{i})P(\hat{h}_{2}(X_{i}) | Y_{i}),...,P(\hat{h}_{M}(X_{i}) | Y_{i})} \\
&=\sum_{i=1}^{N}\mathbf{E}_{(\hat{h}(S_{x}),S_{y})}\log\frac{P(\hat{h}(X_{i}) | Y_{i})}{P(\hat{h}_{1}(X_{i}) | Y_{i})P(\hat{h}_{2}(X_{i}) | Y_{i}),...,P(\hat{h}_{M}(X_{i}) | Y_{i})} \\
&=NI(\hat{h}_{1}(X),\hat{h}_{2}(X),...,\hat{h}_{M}(X)|Y),
\end{aligned}
\label{equ22}
\end{equation}

\begin{equation}
    \sum_{i=1}^{N}I(S_{y},\hat{h}_{i}(S_{x}))+H(S_{y})=N(\sum_{i=1}^{N}I(Y,\hat{h}_{i}(X))+H(Y)).
\label{equ23}
\end{equation}

Combining the Eq. \eqref{equ20},  \eqref{equ21}, \eqref{equ22} and \eqref{equ23}, we can derive the generalization error bound of the fusion model:
\begin{equation}
\begin{aligned}
g(P_{S},P_{W|S})& \leq\frac{1}{N}\sqrt{2\sigma^{2}}\sqrt{NI(f(S);W)} \\
&=\sqrt{2\sigma^{2}}\sqrt{I(\hat{h}_{1}(X),...,\hat{h}_{M}(X))-I(\hat{h}_{1}(X),...,\hat{h}_{M}(X)\mid Y)-H(Y)+\frac{H(W)-H(S_{y},W|\hat{h}(S_{x}))}{N}}.
\end{aligned}
\label{equ24}
\end{equation}

Due to
\begin{equation}
\begin{aligned}
H(W)-H(S_{y},W\mid\hat{h}(S_{x}))\leq H(W)-H(W\mid h(S_{_x}))=I(W;\hat{h}(S_{x})),
\end{aligned}
\label{equ25}
\end{equation}
therefore
\begin{equation}
\begin{aligned}
g\left(P_{S},P_{W|S}\right)&\leq\sqrt{2\sigma^{2}}\sqrt{\underbrace{I(\hat{h}_{1}(X),...,\hat{h}_{M}(X))}_{I_{1}}-\underbrace{I(\hat{h}_{1}(X),...,\hat{h}_{M}(X)|Y)}_{I_{2}}-\underbrace{\sum_{i=1}^{N}I(\hat{h}_{i}(X);Y)}_{I_{3}}+H(Y)+\underbrace{\frac{I(W;\hat{h}(S_{x}))}{n}}_{I_{4}}}.
\end{aligned}
\label{equ26}
\end{equation}

From the derived generalization error bound \citep{Zhang2020RedundancyOH} in part $I_{1}$, it can be observed that as the mutual information between base learners decreases, the model's generalization error bound also decreases.
This indicates the higher independence among the individual base learners, the better performance of the model.
From the perspective of ensemble learning, each base learner can learn different features more independently.
The smaller $I_{2}$ indicates a stronger correlation between the overall features learned by the model and the labels Y, thereby enhancing the model's generalization performance.
$I_{3}$ represents that when there is a higher correlation between the outputs of the base learners and the labels Y, the model's generalization performance improves accordingly.
Therefore, according to the generalization error bound, it can be concluded that in the multi-source data fusion model, greater independence among individual base learners and stronger learning capabilities of each base learner correlate with improved generalization performance of the fused model.
This series of theoretical analysis not only highlights the superior generalization ability of multi-source data fusion model compared to single-source data model, but also provides a robust theoretical basis for subsequent parallel model construction.

\section{Model construction}\label{section: Model construction}
\subsection{The framework for the parallel structure fusion model}\label{section: The framework for the parallel structure fusion model}
Based on the theoretical analysis in Section \ref{section: Theoretical Analysis of Generalization ability in Parallel Fusion Frameworks}, a parallel machine learning framework is proposed to efficiently utilize multi-source data.
This framework integrates multi-source data fusion technology to bolster the model's ability to understand information from various sources, seamlessly merging different information types into a unified representation, thereby addressing the limitations of single-source model.
The framework is capable of integrating complementary information from different data sources to yield a more complete and comprehensive information representation.
It also boosts the system's robustness and prediction stability by providing uniform or similar information.
The model aims to deeply explore the intrinsic relationship among data from various source to optimize predictive performance.
Our objective is to construct a shared space that allows for the integrated expression of multi-source data with differing modalities.

\begin{figure*}[h]
  \centering
   \includegraphics[width=10cm]{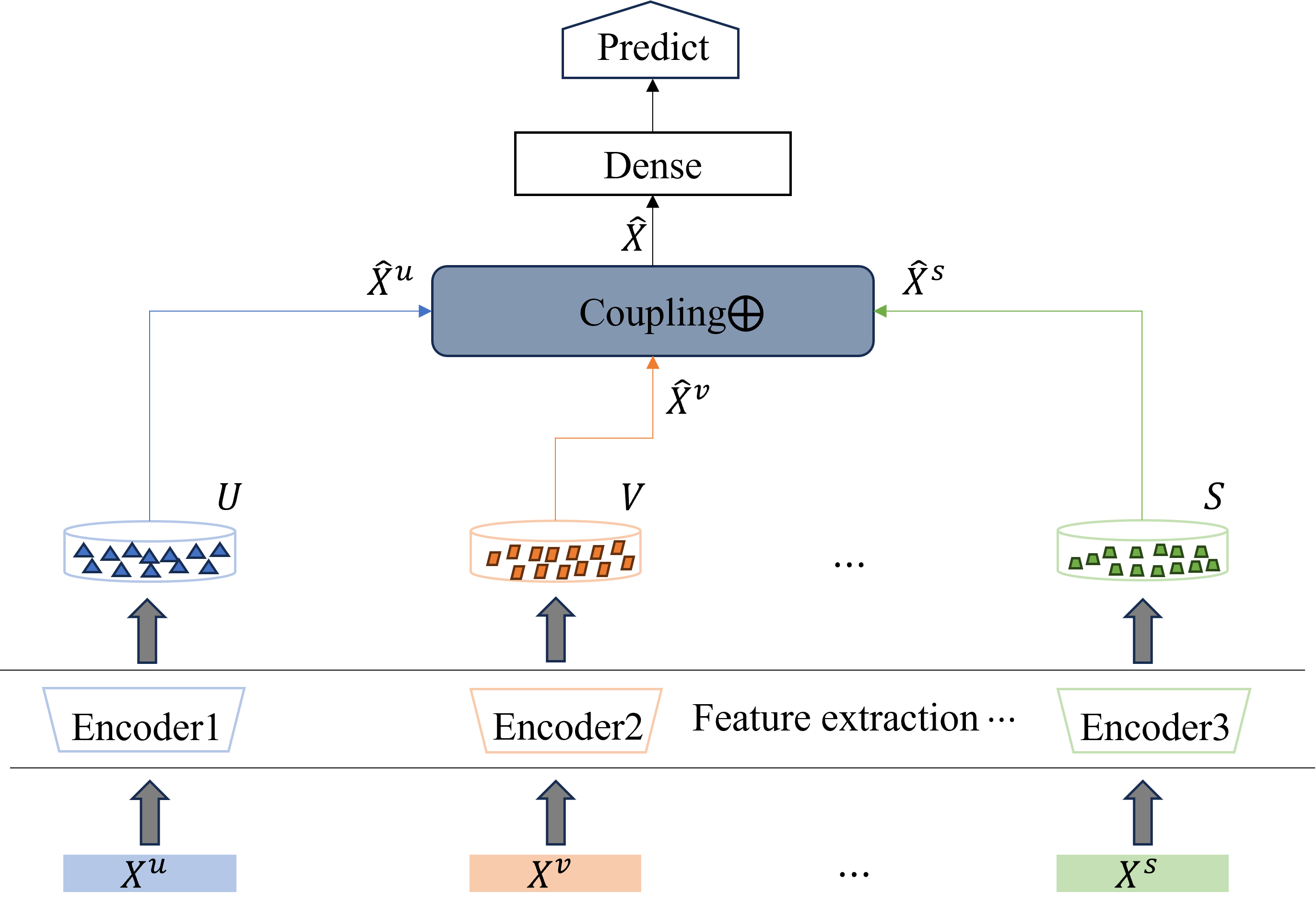}
   \caption{Parallel structure fusion model framework.}
    \label{fig-parallelframework}
\end{figure*}

As shown in Figure \ref{fig-parallelframework}, each data source is first encoded through independent neural networks and subsequently mapped to a shared subspace to extract and fuse features from different data sources, resulting in a unified feature vector.
The construction of this subspace mainly relies on hidden layers, achieved by adding transformed specific vectors.

The features $U,V,\cdots,S$ extracted from different forms $X^{u},X^{v},\cdots,X^{s}$ of sample $X$ are defined as $U=[u_{1},\cdots,u_{{n_{u}}}]^{\top}\in\mathbb{R}^{{n_{u}\times d_{u}}},\quad V=[\nu_{1},\cdots,\nu_{{n_{u}}}]^{\top}\in\mathbb{R}^{{n_{v}\times d_{v}}},\quad\cdots,\quad S=[s_{1},\cdots,s_{{n_{s}}}]^{\top}\in\mathbb{R}^{{n_{s}\times d_{s}}}$, with $n_u,n_v,\cdots,n_s$ denoting the number of feature vectors, and $d_u,d_v,\cdots,d_s$ referring to their dimensionality.
To efficiently integrate the complementarity and redundancy present among different data sources, while considering the properties of feature $U,V,\cdots,S$, the coupling layer is introduced to facilitate feature integration:
\begin{equation}
    \hat{X}^{u}=W_{u}^{\top}U,\\
    \hat{X}^{\nu}=W_{\nu}^{\top}V,\\
    \dots,\\
    \hat{X}^{s}=W_{s}^{\top}S,
\end{equation}
\begin{equation}
    \hat{X}=f(\hat{X}^{u}\oplus\hat{X}^{\nu}\oplus\cdots\oplus\hat{X}^{s}),
\end{equation}
where $\hat{X}$ represents the output nodes of the activation function in the coupling layer, and $W$ denotes the weights of the connection between the specific encoding layer and the coupling layer.
These weights are trainable parameters distinguished by subscripts indicating different data sources.
Defining $g_{i}$ as the loss corresponding to the data source $i$, the learning process of the multi-source data fusion model solves the following optimization problem, aiming to minimize the overall loss:
\begin{equation}
    \min_wg(w)=\frac{1}{n}\sum_{i=1}^ng_i(w),
\end{equation}
\begin{equation}
    g_i(w)=\mathbf{E}_{\left(X_i^j,Y_i^j\right)\in\mathcal{D}_i}[l_i(w;X_i^j,Y_i^j)],
\end{equation}
where L is the error of the model in predicting the true label Y.

\subsection{The proposed parallel cnn-gru attention model}\label{section: The proposed parallel cnn-gru attention model}
Based on the theoretical analysis in Section \ref{section: Theoretical Analysis of Generalization ability in Parallel Fusion Frameworks}, this study introduces a novel parallel CNN-GRU attention model for power load forecasting.
The proposed model combines the capabilities of CNN and GRU to extract both spatial and temporal features from the multivariate time series data.
The CNN network is employed to capture spatial information, while the GRU layer is utilized to extract and learn temporal features.
Additionally, an attention mechanism is incorporated to emphasize the importance of key features in the forecasting process.
Our research focuses on utilizing multivariate time series data that contains abundant spatio-temporal information.
By leveraging the parallel CNN-GRU attention model, our objective is to enhance the accuracy and efficiency of load forecasting.
This improvement in forecasting accuracy will provide valuable data support for decision-making and planning in relevant fields.
The integration of spatial and temporal information through our proposed model will enable more informed and effective decision-making processes.
\begin{figure*}[h]
  \centering
   \includegraphics[width=15.5cm]{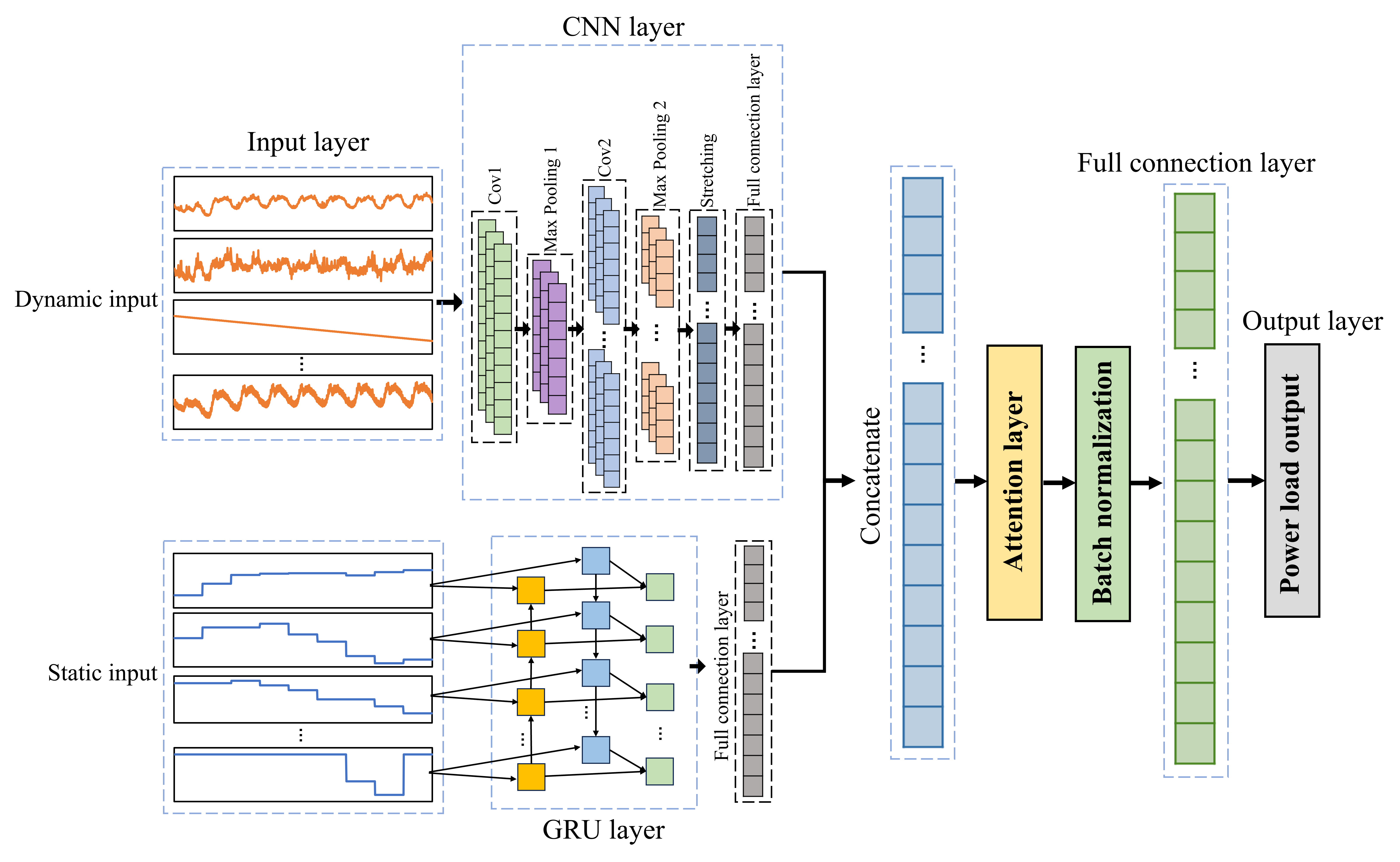}
   \caption{The parallel cnn-gru attention model framework.}
    \label{fig1}
\end{figure*}

The frame structure of the proposed model is depicted in Figure \ref{fig1}.
Firstly, the multivariate time series data associated with the power load, which is composed of the filtered features, is transformed into the multivariate input array.
Subsequently, the data is fed into the CNN and GRU modules for training.
The CNN module comprises two convolutional layers, two pooling layers, and a fully connected layer.
The convolution layer 1 and 2 are designed as one-dimensional convolution layers, and the specific parameter settings are listed in the later experimental section.
The two pooling layers employ the principle of maximum pooling for the pooling operation.
The feature data extracted by the convolutional layer is transformed into a one-dimensional vector and subsequently processed by the fully connected layer.
Meanwhile, the bidirectional GRU model is employed to train the features of the dynamic part, and the output is likewise transformed into a one-dimensional vector and processed by the fully connected layer.
Subsequently, each output of the CNN and GRU modules is cascaded and fed into the attention layer.
The attention layer calculates the attention score of the vector composed of significant spatio-temporal features, and the resulting score is inputted into the final fully connected layer.
Finally, through batch normalization and two fully connected layers, we get the prediction results of power load.
During the iterative process of the model, the mean square error loss function is utilized to train the model.

\begin{figure*}[htb]
  \centering
   \includegraphics[width=15.5cm]{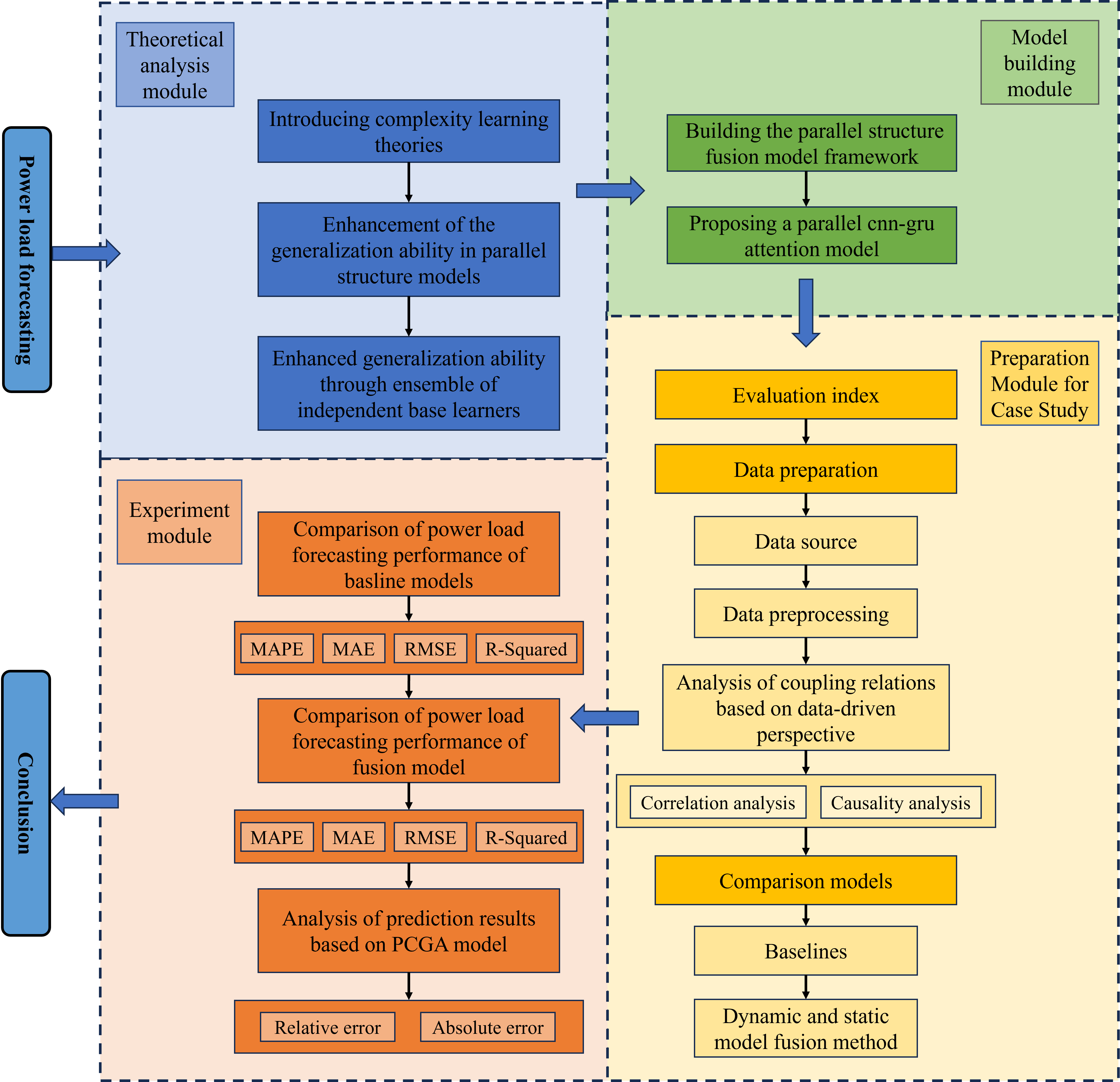}
   \caption{Work flow chart.}
    \label{fig-workflow}
\end{figure*}

The structure of the CNN-GRU Attention model adheres to the parallel form model derived in Section \ref{section: Theoretical Analysis of Generalization ability in Parallel Fusion Frameworks}.
This structural design significantly enhances the model's generalization capabilities compared to traditional single and serial models.
Therefore, the CNN-GRU Attention model excels in extracting static spatial factors and  the long-term correlation of dynamic time series data.
This multi-level and multi-dimensional approach allows the model to better capture complex patterns and long-term dependencies in data, thereby demonstrating superior performance in power load forecasting.
Figure \ref{fig-workflow} illustrates the complete workflow developed in this paper for tackling power load issue.

\section{Preparation of case study}\label{section: Preparation of case study}
\subsection{Evaluation index}\label{section: Evaluation index}
In this experiment, we used mean absolute percentage error (MAPE) \citep{Myttenaere2016MeanAP}, mean absolute error (MAE) \citep{Qi2020OnMA}, root mean square error (RMSE) \citep{Karunasingha2021RootMS} and R2 as performance indicators to evaluate each model in load forecasting tasks.
These indicators can provide a quantitative assessment of the prediction error and model fitting ability.
The larger MAPE, MAE and RMSE values indicate that the prediction error is larger, while the larger R2 value indicates that the fitting effect of the model is better.
These indicators are defined as follows:
\begin{equation}
\begin{aligned}
MAPE=\frac{1}{n}\sum_{i=1}^n\left|\frac{\hat{y}_i-y_i}{y_i}\right|\times100\%
\end{aligned}
\end{equation}
\begin{equation}
MAE=\frac1n\sum_{i=1}^n\mid\hat{y}_i-y_i\mid
\end{equation}
\begin{equation}
RMSE=\sqrt{\frac1n\sum_{i=1}^n(\hat{y}_i-y_i)^2}
\end{equation}
\begin{equation}
\begin{aligned}
R^2&=1-\frac{\sum(y_i-\hat{y}_i)^2}{\sum_{i=1}^n(y_i-\bar{y})^2}
\end{aligned}
\end{equation}

Here, $n$ represents the number of data points, $y_i$ represents the actual heat load of the $i$-th data point, $\hat{y}_{i}$ represents the predicted heat load of the $i$-th data point, and $\bar{y}$ represents the average heat load.
By utilizing these evaluation indicators, we can objectively assess the performance of various models in load forecasting tasks and provide guidance and a foundation for model selection and optimization.
Utilizing these indicators enables a quantitative assessment of the model's prediction accuracy and error level, facilitates comprehension of the model's fit with the actual data, and enhances the capability to enhance and optimize the model.

\subsection{Data preparation}\label{section: Data preparation}
\subsubsection{Data source}\label{section: Data source}
The original data used in this study for power load forecasting includes 128,156 power system load data from January 1, 2018 to August 31, 2021, which are sampled at intervals of 15 minutes.
It also includes 1,345 daily sampling meteorological condition factor data.
The collected data can provide sufficient time series information and provide a necessary basis for the establishment and evaluation of power load forecasting models.
\subsubsection{Data preprocessing}\label{section: Data preprocessing}
For missing values in the data, considering the periodic dependence of time series data, we use the average value of the 7-day load data preceding and following the missing data to estimate the missing values.
Outliers are categorized as anomalies resulting from holidays and anomalies arising from uncontrollable factors.
To address outliers caused by holidays, we incorporate a binary feature indicating whether it is a holiday as an input.
Outliers resulting from uncontrollable factors are treated as missing values and subsequently interpolated and corrected.

To improve the prediction accuracy of power load values, we construct additional factors related to power load.
The power load forecasting dataset is composed of historical load-related data, date factors, and meteorological factors, which are summarized in Table \ref{Table1}.

The historical load data includes the original load sequence, as well as the average daily load, trend, seasonal, and residual sequences obtained through time series decomposition of the original load sequence.
In the date factor, different dates are represented using positive integers in ascending order. For example, spring, summer, autumn, and winter can be represented as 1, 2, 3, and 4 respectively. Similarly, January to December can be represented as 1 to 12, and so on.
The meteorological factors include maximum temperature, minimum temperature, weather conditions, daytime wind force, daytime wind direction, nighttime wind force, and nighttime wind direction. These factors provide important information for analyzing and forecasting weather patterns.
Among this factors, the weather condition is a character type category variable. To facilitate analysis, we will replace them with numerical values in ascending order based on the magnitude of weather conditions, ranging from sunny to rainstorm.
The wind force characteristics can be mapped to sequential values based on the magnitude of the wind force, while the wind direction has no size relationship. Therefore, it is necessary to perform one-hot coding and convert it into a data form that the model can handle.
\setlength{\tabcolsep}{8pt}
\begin{table*}[htbp]
\caption{Categorization and description of the utilized data.}
\centering
\resizebox{\textwidth}{!}{
\begin{tabular}{lll}
\toprule
Category & Explanation & Description\\
\hline
\multirow{3}{*}{Historical load factors}
& History Load & Original load sequence\\
& Average load & The average load at 96 sampling time points per day\\
& Decomposition results & Seasonal and trend decomposition results\\
\hline
Data factors &  & Day, week, month, season, whether the weekend, whether the holidays\\
Meteorological factor & & Temperature, weather conditions, wind force, wind direction\\
\bottomrule
\end{tabular}
}
\label{Table1}
\end{table*}

\begin{figure*}[htb]
  \centering
   \includegraphics[width=15.5cm]{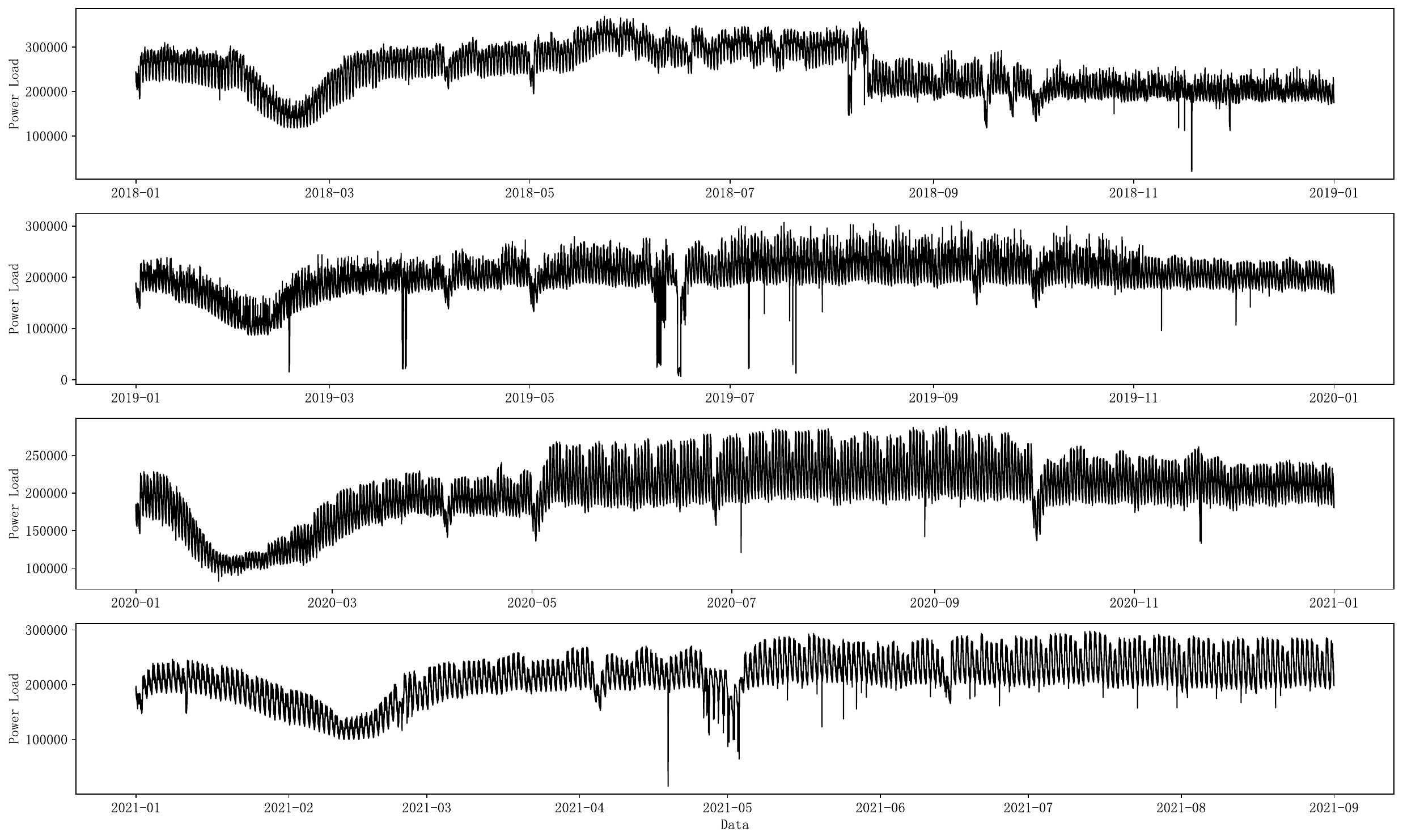}
   \caption{Preprocessed power load historical data.}
    \label{fig2}
\end{figure*}
\begin{figure*}[htb]
  \centering
   \includegraphics[width=14cm]{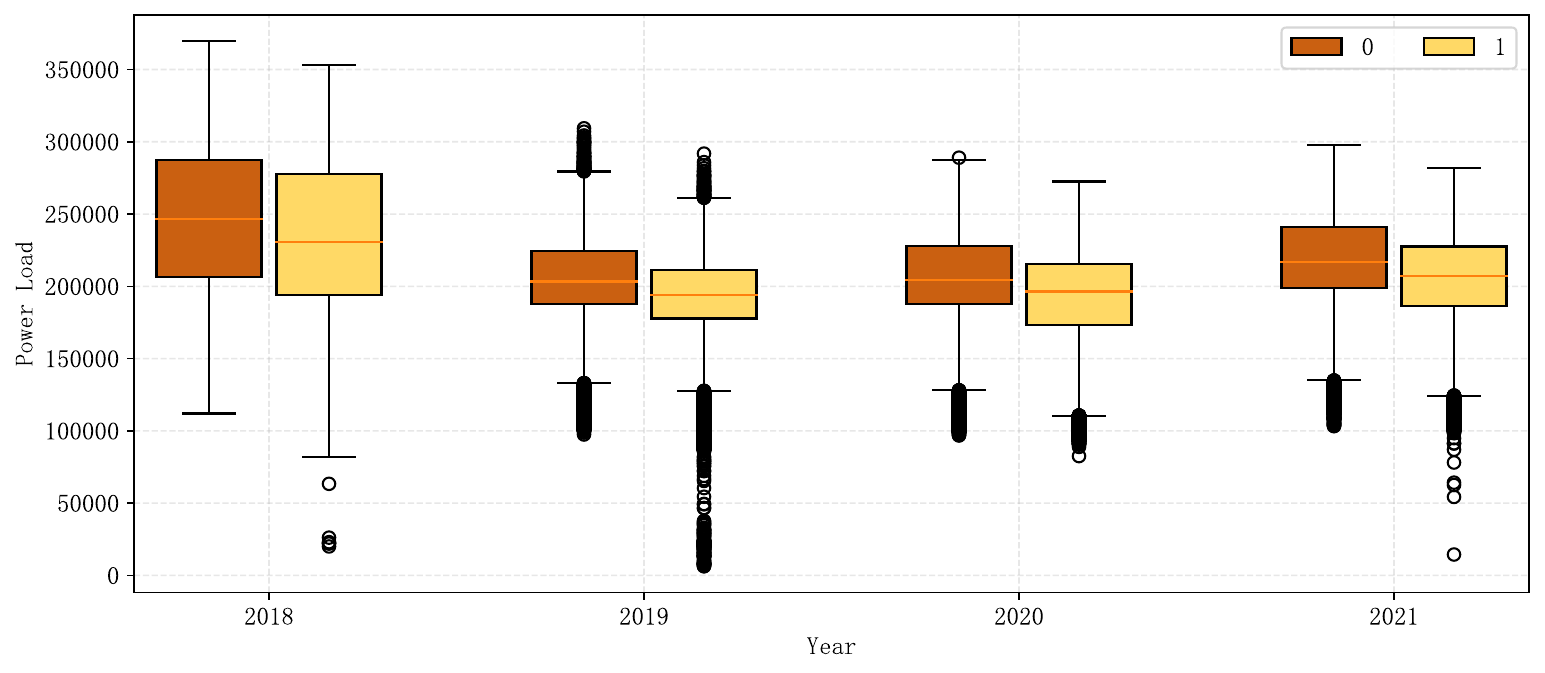}
   \caption{Power load patterns of holidays and non-holidays (label 0 represents non-holidays, label 1 represents holidays).}
    \label{fig3}
\end{figure*}
\begin{figure*}[htb]
  \centering
   \includegraphics[width=14cm]{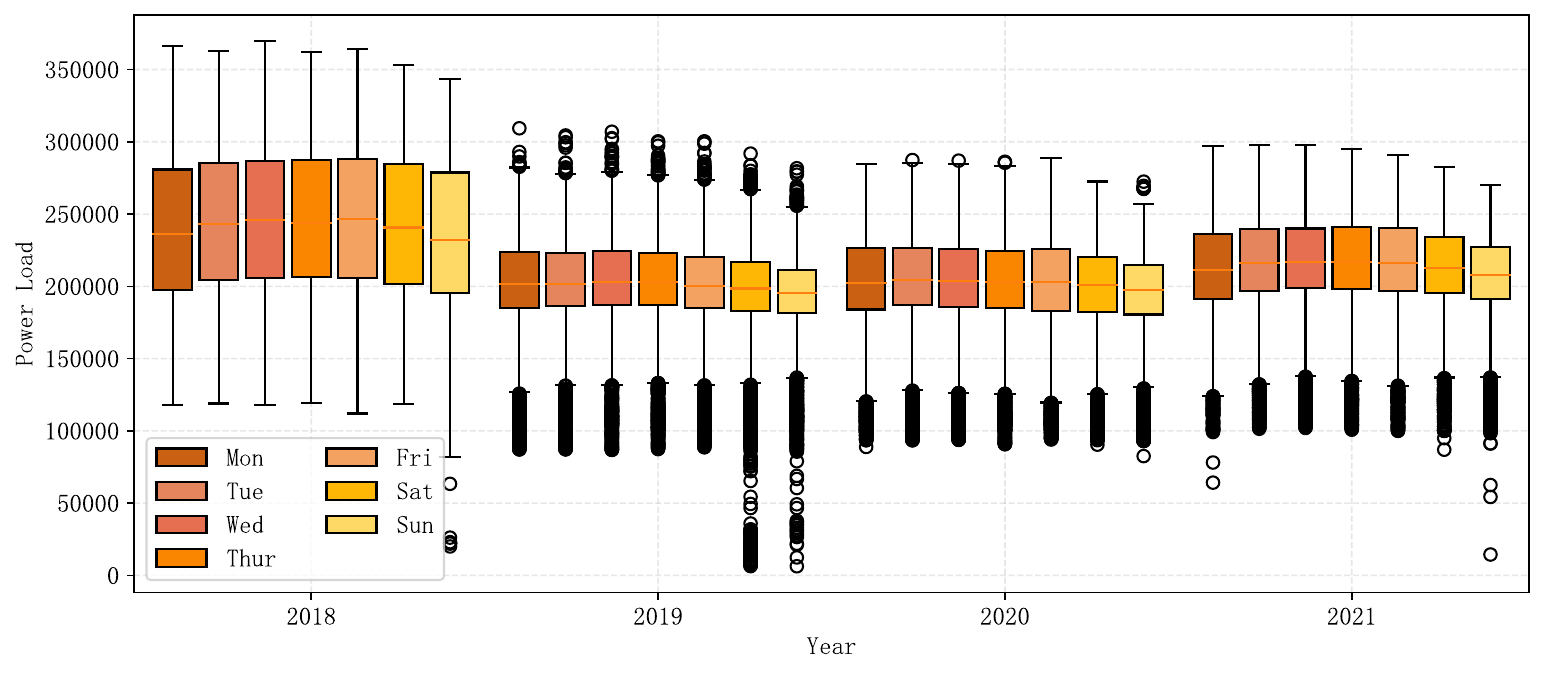}
   \caption{Power load patterns from Monday to Sunday.}
    \label{fig4}
\end{figure*}

\subsubsection{Analysis of coupling relations based on data-driven perspective}\label{section: Analysis of coupling relations based on data-driven perspective}
According to the data preprocessing process described in Section \ref{section: Data preprocessing}, we process the collected data with outliers and missing values.
The processed power load time series data is shown in Figure \ref{fig2}.
It can be clearly observed from the figure that there are obvious differences between the heat load patterns in spring, summer, autumn and winter, showing a clear seasonal trend.
Next, we analyze the heat load patterns on holidays and non-holidays, as shown in Figure \ref{fig3}.
It can be clearly seen from the figure that there are obvious differences in power load patterns between non-holidays and holidays.
In general, the power load level of non-holidays is higher than that of holidays.
In addition, we also analyze the power load pattern from Monday to Sunday, as shown in Figure \ref{fig4}.
It can be observed that during the week, the power load levels on Saturdays and Sundays are generally lower than those on weekdays, and the difference on Sundays is particularly significant.
These observations provide important clues for further analysis of seasonal and periodic changes in power load, and help us to understand the changing patterns and trends of power load more deeply.

Based on our observation and trend analysis of power load change patterns, we have comprehensively summarized the factors affecting power load.
In order to improve the accuracy of power load forecasting, we introduce a series of influencing factors related to power load forecasting based on the original load time series data.
These factors include date-related factors (covering information on days, weeks, months, seasons, whether weekends, whether holidays, etc.), and trend factors (including seasonal and trend decomposition results of the original sequence, and the average load of the day on which the current load is located).
All these factors are shown in detail in Table \ref{Table2}.
Comprehensively considering these factors will help us to predict power load more accurately and better understand the various dynamic and static factors behind power load fluctuations.
\setlength{\tabcolsep}{8pt}
\begin{table*}[htbp]
\caption{All relevant factors and their corresponding numbers of pre-processed power load forecasting.}
\centering
\resizebox{\textwidth}{!}{
\begin{tabular}{llll}
\toprule
1 & Weather conditions 1 & 21 & \textbf{Year}\\
\hline
2 & Weather conditions 2 & 22 & \textbf{Quarter}\\
\hline
3 & Daytime wind direction: northeast wind & 23 & \textbf{Month}\\
\hline
4 & Daytime wind direction: southeast wind & 24 & Day\\
\hline
5 & Daytime wind direction: east wind & 25 & \textbf{Hour}\\
\hline
6 & Daytime wind direction: north wind & 26 & The day of the week\\
\hline
7 & Daytime wind direction: south wind & 27 & Is weekday?\\
\hline
8 & Daytime wind direction: no continuous wind direction & 28 & Is holiday?\\
\hline
9 & Daytime wind direction: Southwest wind & 29 & \textbf{Holiday type}\\
\hline
10 & Night wind direction: northeast wind & 30 & Holiday: Dragon Boat Festival\\
\hline
11 & Night wind direction: southeast wind & 31 & Holiday: Labour Day\\
\hline
12 & Night wind direction: east wind & 32 & Holiday: Mid-autumn Festival\\
\hline
13 & Night wind direction: north wind & 33 & Holiday: National Day\\
\hline
14 & Night wind direction: south wind & 34 & Holiday: New Year's Day\\
\hline
15 & Night wind direction: no continuous wind direction & 35 & \textbf{Holiday: Spring Festival}\\
\hline
16 & Night wind direction: southwest wind & 36 & Holiday: Tomb-sweeping Day\\
\hline
17 & \textbf{Daytime wind} & 37 & \textbf{The average total active power of the day}\\
\hline
18 & \textbf{Night wind} & 38 & \textbf{Seasonal decomposition sequence}\\
\hline
19 & \textbf{Maximum temperature} & 39 & \textbf{Trend decomposition sequence}\\
\hline
20 & \textbf{Minimum temperature} & 40 & \textbf{Residual sequence}\\
\bottomrule
\end{tabular}
}
\label{Table2}
\end{table*}

\vspace{8pt}
a) Correlation analysis

We will use Pearson correlation coefficient \citep{Jiang2021MediumlongTL}, Spearman correlation coefficient \citep{Chen2023MultifeatureSP} and Kendall correlation coefficient \citep{Blk2023ImplementationOD} to analyze the correlation between power load and various influencing factors.
Pearson correlation coefficient is used to measure the degree of linear relationship between random variables ${X}$ and ${Y}$, which is suitable for continuous variables and satisfies the assumption of normal distribution (i.e., implicit Gaussian distribution) \citep{Xu2018DependentEC}.
The Spearman correlation coefficient is used to describe the monotonic relationship between random variables ${X}$ and ${Y}$, which is not affected by outliers and is suitable for the case of nonlinear relationships \citep{Deebani2020MonteCE}.
The Kendall correlation coefficient, like the Spearman correlation coefficient, is a rank correlation coefficient.
However, it assesses the correlation strength between sample data pairs based on their relationship, rather than relying on rank differences \citep{Huang2022InfluenceFC}.
Through these three correlation coefficients, we can fully understand the complex correlation between power load and various influencing factors.

Additionally, we also use copula entropy (CE).
The concept of copula entropy is defined by Copula density function, which is essentially a form of Shannon entropy \citep{Mortezanejad2019JointDD}.
CE is a more advanced correlation measure that has significant advantages over Pearson correlation coefficients. Unlike Pearson correlation, CE does not assume linearity and Gaussianity, and it can handle multivariate correlations.
In fact, CE measures statistical independence, which is a broader concept than correlation. When two variables are statistically independent, the CE value is 0.
CE also has monotonic transformation invariance, which is equivalent to the correlation coefficient in the case of Gaussian distribution \citep{Sun2021NonparametriccopulaentropyAN}.

\vspace{8pt}
a) Causality analysis

Granger causality test is a statistical method of hypothesis testing. It tests whether a set of time series ${x}$ is the cause of another set of time series ${y}$, and uses the historical data of ${x}$ to improve the ability to predict ${y}$ variables \citep{Chopra2018StatisticalTF}.

Let ${x}$ and ${y}$ be generalized stationary sequences. Firstly, the ${p}$-order autoregressive model of ${y}$ is established, and then the lag period of ${x}$ is introduced to establish the augmented regression model, i.e,
\begin{equation}
\begin{cases}
y_t=c_1+\sum_{i=1}^p\alpha_{11}^{(i)}y_{t-i}+\sum_{i=1}^p\alpha_{12}^{(i)}x_{t-i}+u_{1t}\\x_t=c_2+\sum_{i=1}^p\alpha_{21}^{(i)}y_{t-i}+\sum_{i=1}^p\alpha_{22}^{(i)}x_{t-i}+u_{2t}
\end{cases}
\end{equation}
Where $c_1$ and $c_2$ are constants, $\alpha_{st}^{(i)}(s,t=1,2)$ is the ${i}$-th lag autoregressive coefficient, ${u}_{1t}$ and ${u}_{2t}$ are errors, $x_{t-i}$ and $y_{t-i}\left(i=1,2,\cdot\cdot\cdot,p\right)$ are the observed values of the ${i}$-th period of ${x}$ and ${y}$, respectively.

Taking $y_t$ as an example, if the autoregressive coefficient satisfies $\alpha_{12}^{(1)}=\alpha_{12}^{(2)}=\cdotp\cdotp\cdotp=\alpha_{12}^{(p)}=0$, then there is no Granger causality of $x_t\to y_t$. The F test is usually used to determine whether there is Granger causality \citep{Jafari2023MVARAC}.

\begin{figure*}[htb]
  \centering
   \includegraphics[width=15.5cm]{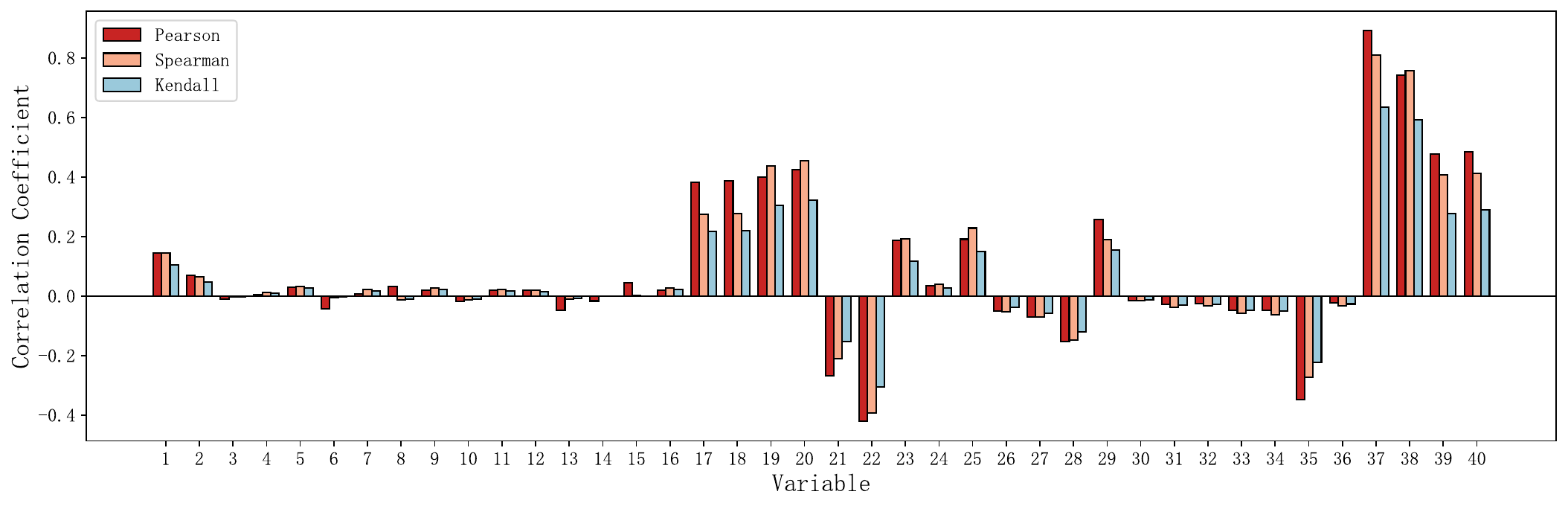}
   \caption{The correlation coefficient between each input factor and the load time series to be predicted.}
    \label{fig5}
\end{figure*}

Furthermore, in order to reduce the data dimension, improve the prediction efficiency of the model and ensure that the input factors can have a positive impact on the prediction results, we performed a feature selection operation.
Firstly, we encode the categorical variables before feature selection.
Due to the large number of encoded features, we use numbering to represent each feature.
Table \ref{Table2} lists the names and corresponding numbers of each feature.
Next, we first filter these features based on correlation.
Figure \ref{fig5} shows the results of correlation analysis, including Pearson correlation coefficient, Spearman correlation coefficient and Kendall correlation coefficient between various input factors and the load time series to be predicted.
It is evident from the figure that the correlation coefficients of feature numbers 17,18,19,20,21,22,23,25,29,35,37,38,39, and 40 are significantly higher than those of other features.
In particular, the correlation coefficients between the features numbered 37 and 38 and the load series exceed 0.7, indicating that there is a strong correlation between these two features and the load series.
The absolute value of the correlation coefficient between other features and the load sequence is between 0.2 and 0.5, indicating that they have a moderate correlation with the load sequence.
\begin{figure*}[htb]
  \centering
   \includegraphics[width=15.5cm]{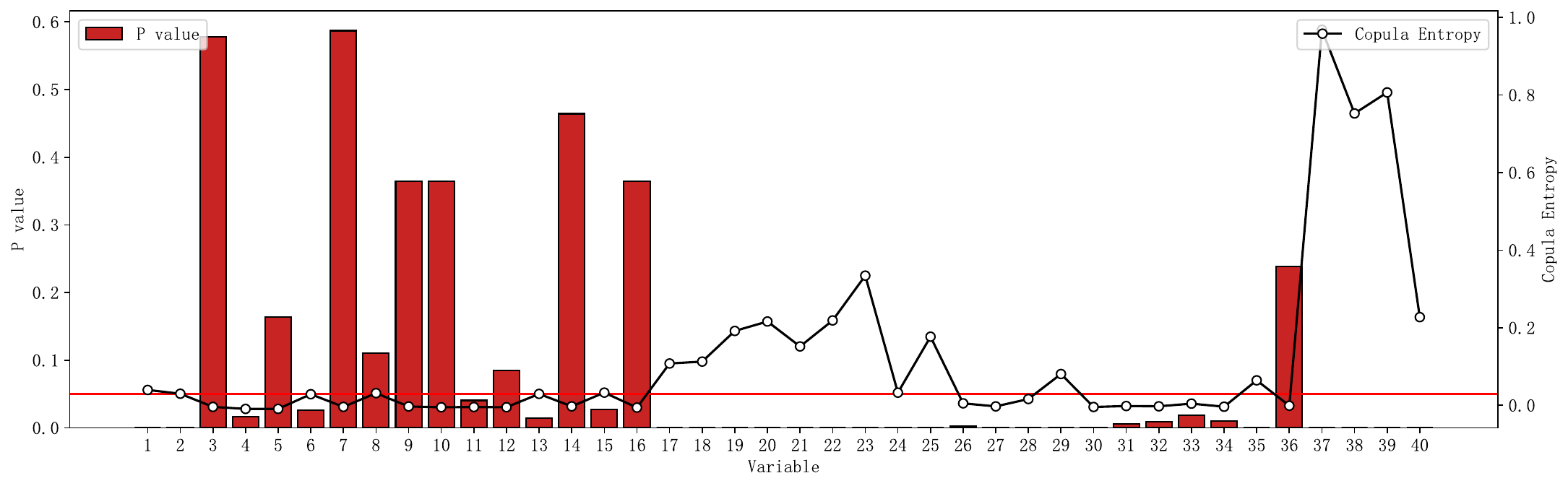}
   \caption{Copula entropy and Granger causality test results.}
    \label{fig6}
\end{figure*}

Figure \ref{fig6} shows the Copula entropy between each feature and the load sequence, as well as the $P$-values of the $F$-test to determine whether there is Granger causality.
Copula entropy not only measures the statistical dependence between variables and responses, but also reflects the information transfer or energy exchange in the underlying system.
Compared with the traditional correlation coefficient, the Copula entropy has richer information, because it considers the nonlinear relationship between variables and the potential joint distribution structure.
Granger causality test is a method used to test whether a set of time series $x$ has a causal effect on another set of time series $y$.
It uses the historical data of $x$ to improve the predictive ability for the variable $y$.
In the Figure \ref{fig6}, we draw a red line representing a $P$-value of 0.05.
Features with $P$-value lower than the red line indicate that they satisfy the Granger causality test, that is, they have a predictive effect on the change of the load sequence.
It can be clearly observed from the figure that the features numbered 17,18,19,20,21,22,23,25,29,35,37,38,39, and 40 not only satisfy the Granger causality test, but also have significantly higher Copula entropy values are than other features.
This further verifies that these features play an important role in power load forecasting and are therefore retained as input factors for subsequent model tests.

\subsection{Comparison models}\label{section: Comparison models}
\subsubsection{Baselines}\label{section: Baselines}
This section introduces the baseline model used before constructing the parallel prediction model.
BP neural network and CNN convolutional neural network are used as baseline models for information extraction of static features.
BP neural network is a multi-layer feedforward neural network trained according to the error back propagation algorithm \citep{Li2009TheIT}.
CNN convolutional neural network is a feedforward neural network with convolution calculation and deep structure \citep{Wu2020ApplicationOI}.
LSTM, BiLSTM and GRU are used as baseline models for information extraction of dynamic features.
LSTM is an algorithm to improve the gradient vanishing problem of RNN.
It introduces three gates on ordinary RNN to avoid long-term dependence problem \citep{Liu2021SimulationOD}.
BiLSTM combines the forward-propagating LSTM and backward-propagating LSTM, enabling the model to capture information from both past and future contexts.
GRU, a variant of LSTM, has a simpler structure with one less gate function.
As a result, GRU tends to have faster training speeds compared to LSTM \citep{Chen2021StateOC}.

\subsubsection{Dynamic and static model fusion method}\label{section: Dynamic and static model fusion method}
Two fusion approaches are employed to combine the baseline models for extracting static and dynamic feature information.
One approach involves a serial model structure, where all the data is first input into the baseline model for extracting static feature information, and the resulting output is subsequently fed into the baseline model for extracting dynamic feature information.
The other approach employs a parallel model structure, where the static data is input into the baseline model for extracting static feature information, while the dynamic data is simultaneously input into the baseline model for extracting dynamic feature information.
The outputs of both parts of the model are connected through a fully connected layer to facilitate information fusion.
These fusion approaches enable the integration of both static and dynamic features, enhancing the overall performance of the information extraction system.

Because the serial structure first extracts the spatial features of the data through the BP layer or the CNN layer, that is, the static features, and then uses the extracted features as the input data of the dynamic baseline model layer (LSTM layer or BiLSTM layer or GRU layer), the dynamic model layer cannot extract the time series dynamic change features from the original time series data.
The parallel structure can synchronously extract the static and dynamic features of the original data, which shows its advantages.
In this study, two baseline models for information extraction of static features and three baseline models for information extraction of dynamic features are combined in pairs.
The subsequent experimental section will compare models constructed using a serial structure with those developed through a parallel structure.
This comparison aims to confirm that the parallel-structured model exhibits enhanced performance.
\begin{figure*}[ht]
  \centering
  \subfigure[]{
    \label{Fig7-a}
    \includegraphics[width=7cm]{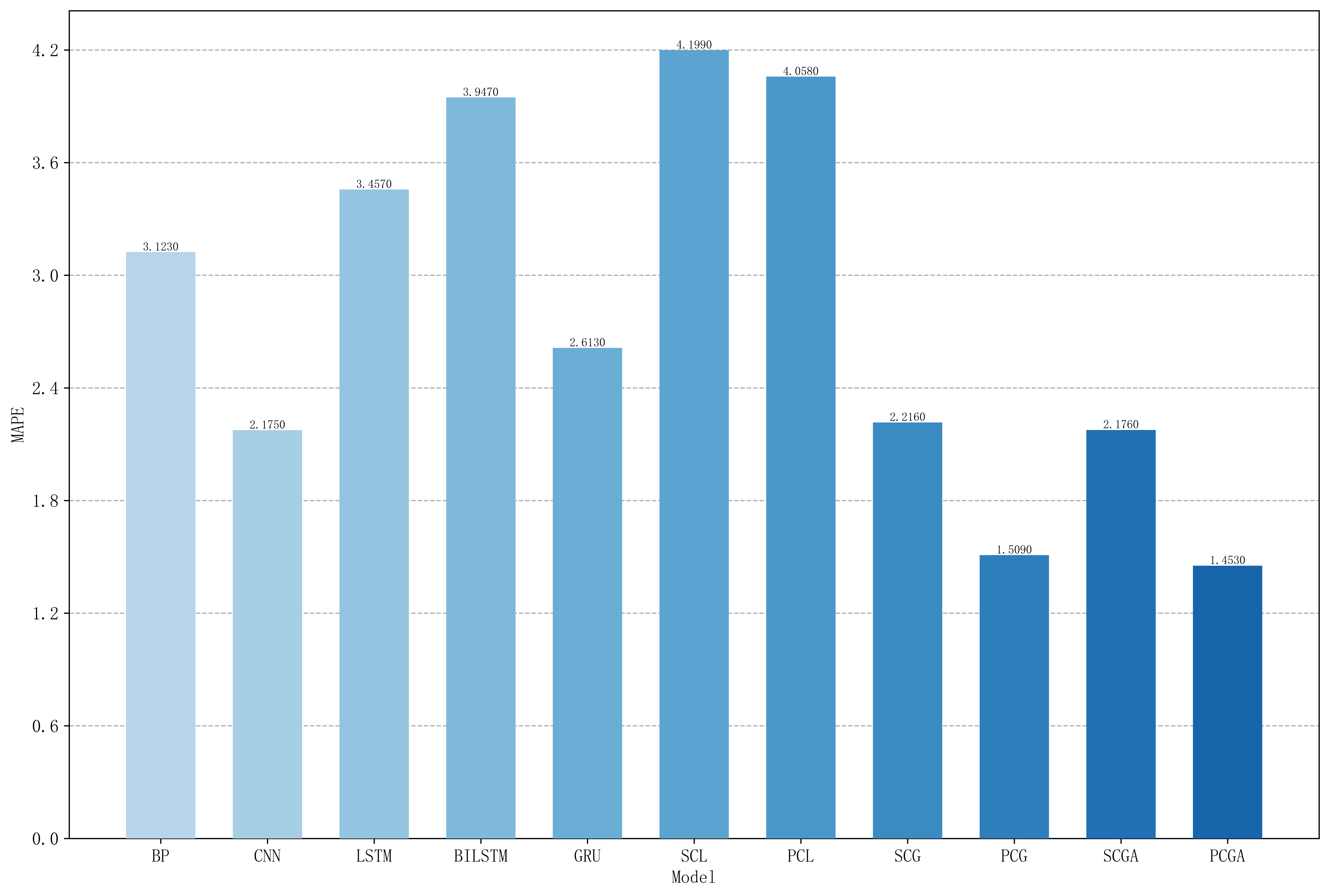}}
    \hspace{-0.1in}
  \subfigure[]{
    \label{Fig7-b}
    \includegraphics[width=7cm]{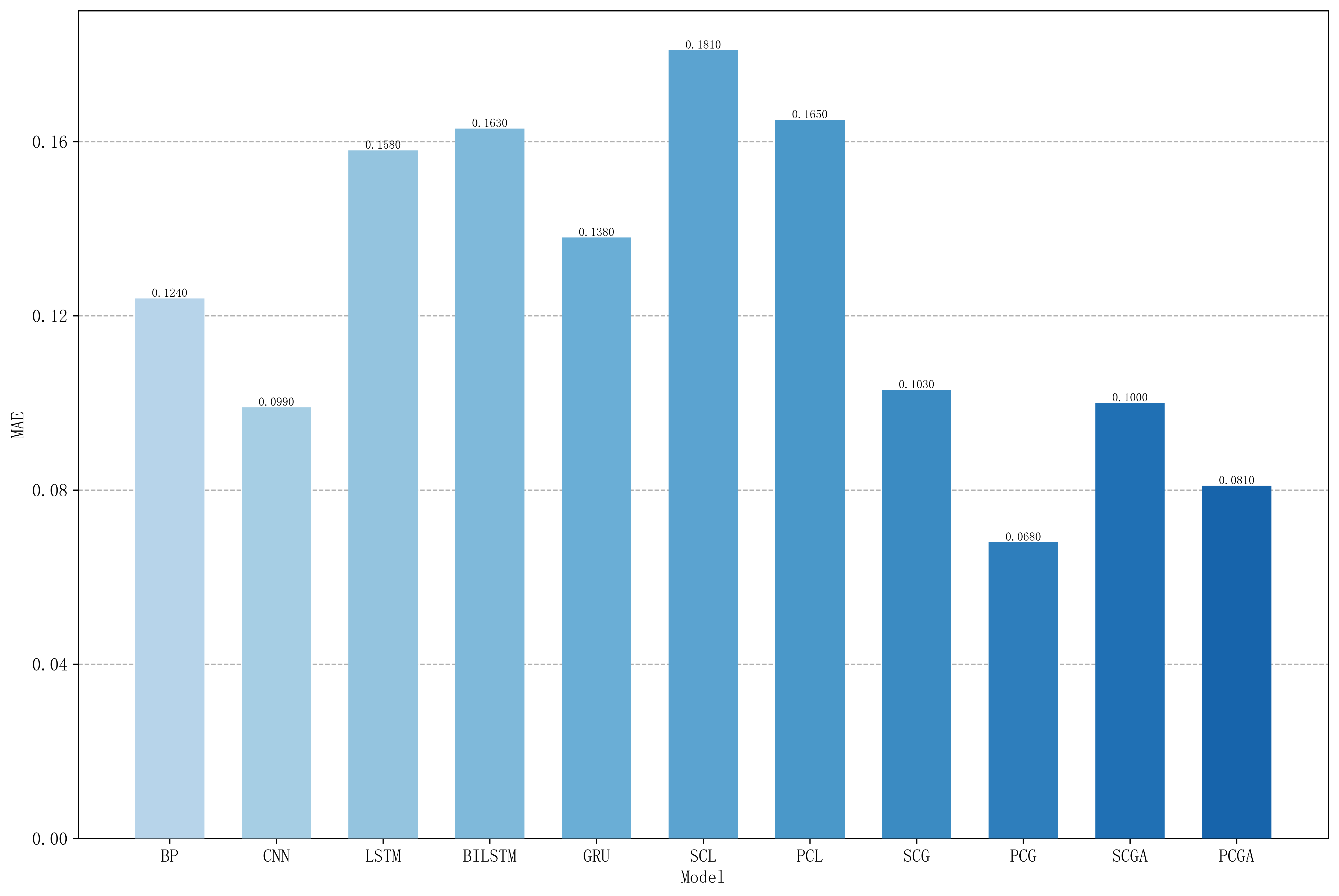}}
    \hspace{-0.1in}
  \subfigure[]{
    \label{Fig7-c}
    \includegraphics[width=7cm]{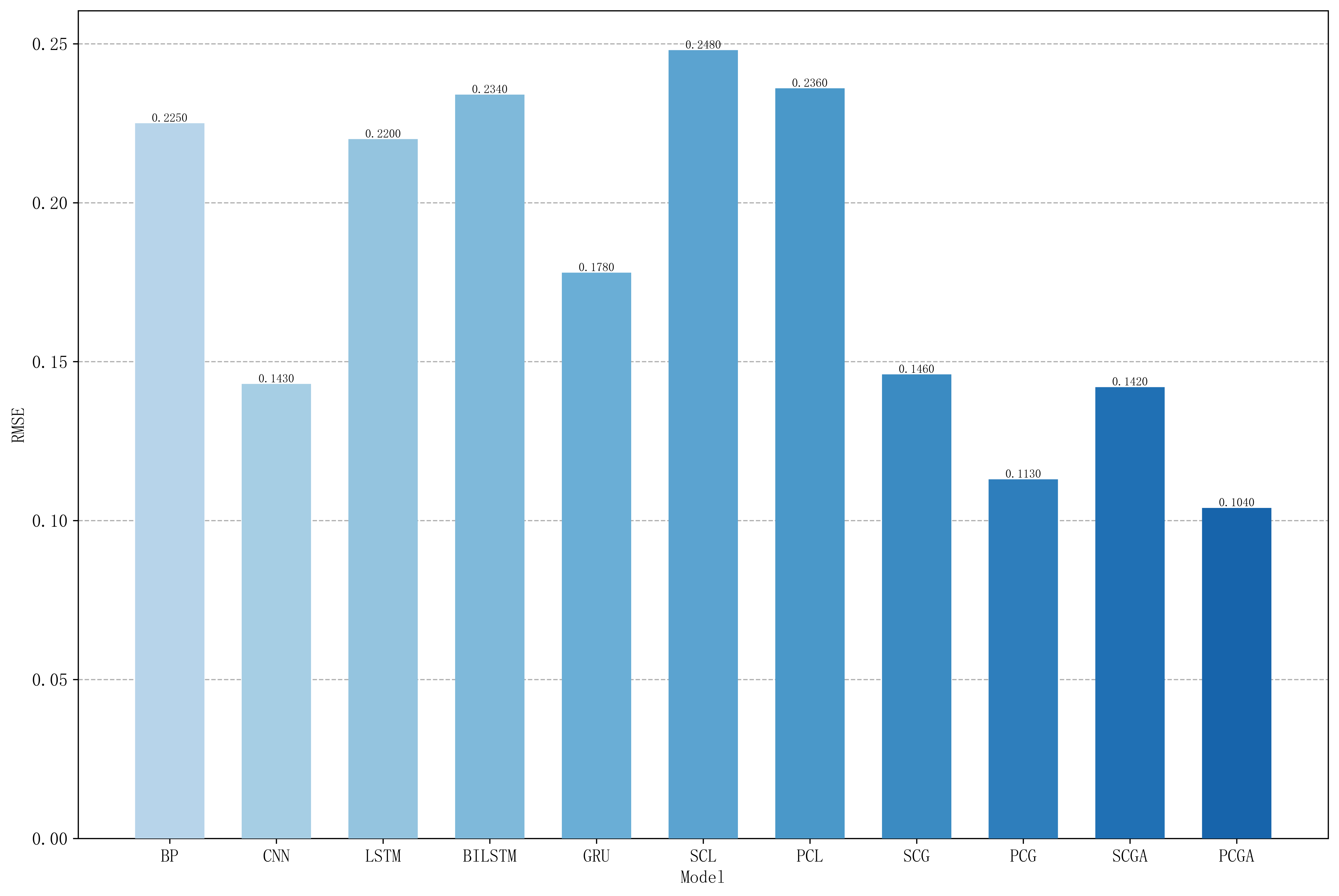}}
    \hspace{-0.1in}
  \subfigure[]{
    \label{Fig7-d}
    \includegraphics[width=7cm]{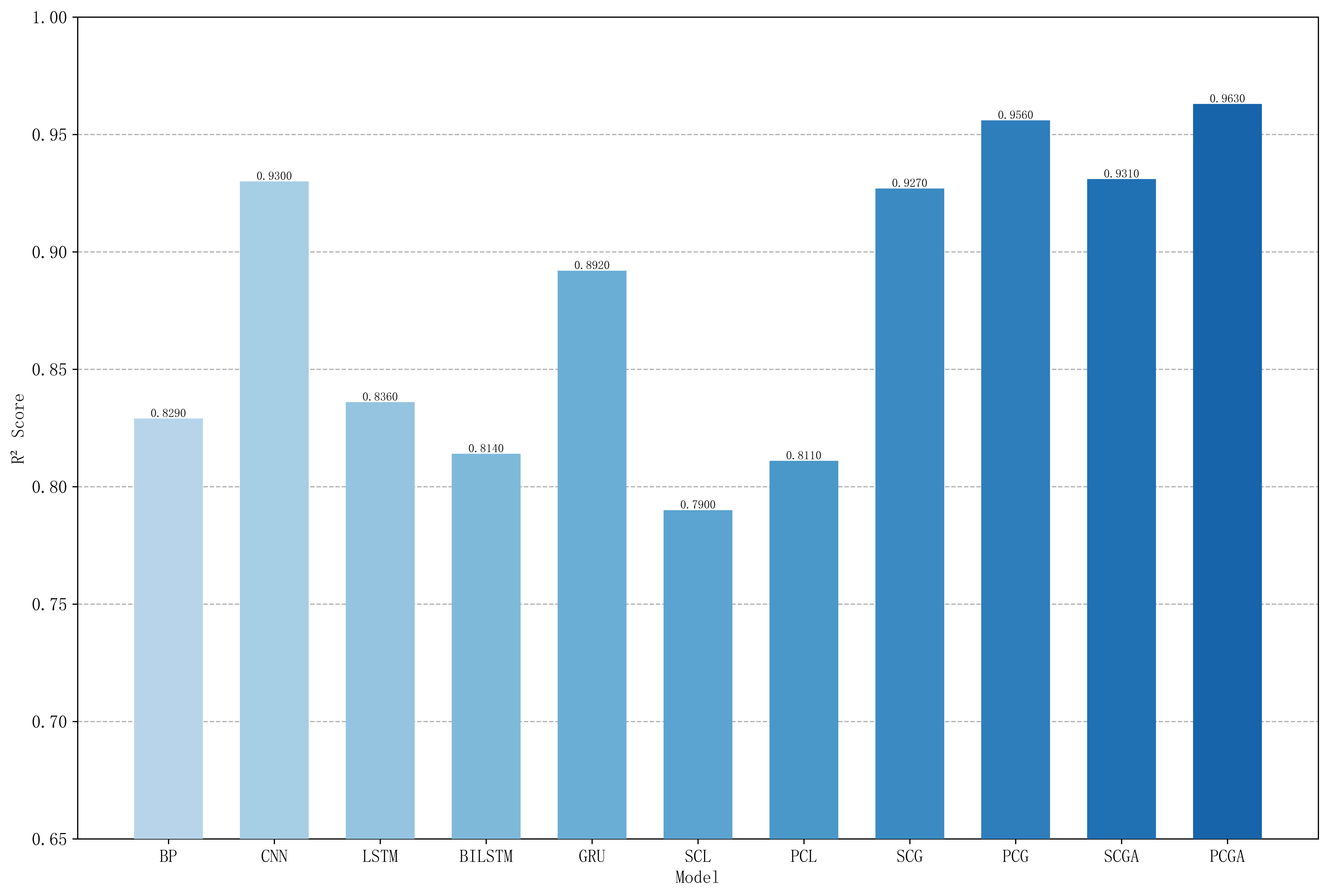}}
    \caption{Comparison of MAPE, MAE, RMSE, $R^2$ score of each model.}
  \label{fig7}
\end{figure*}
\section{Experimental result}\label{section: Experimental result}
\subsection{Comparison of power load forecasting performance of basline models}\label{section: Comparison of power load forecasting performance of basline models}
\setlength{\tabcolsep}{8pt}
\begin{table*}[htbp]
\caption{Model parameter settings.}
\centering
\resizebox{\textwidth}{!}{
\begin{tabular}{llllll}
\toprule
Hyperparameter & BP & CNN & LSTM & BILSTM & GRU\\
\hline
Epoch & 50 & 50 & 50 & 50 & 50\\
\hline
Batch size & 64 & 64 & 64 & 64 & 64\\
\hline
Learning rate & $10^{-4}$ & $10^{-4}$ & $10^{-4}$ & $10^{-4}$ & $10^{-4}$\\
\hline
Dropout rate & 0.3 & 0.3 & 0.3 & 0.3 & 0.3\\
\hline
decay rate & 0.5 & 0.5 & 0.5 & 0.5 & 0.5\\
\hline
Activation function & ReLU & ReLU & ReLU & ReLU & ReLU\\
\hline
Hidden layers & 2 & \textbackslash & 2 & 2 & 2\\
\hline
Number of hidden layer nodes 1 & 150 & \textbackslash & 150 & 150 & 150\\
\hline
Number of hidden layer nodes 2 & 150 & \textbackslash & 150 & 150 & 150\\
\hline
Bidirectional & \textbackslash & \textbackslash & False & True & True\\
\hline
Convolution layers & \textbackslash & 2 & \textbackslash & \textbackslash & \textbackslash\\
\hline
Number of channels 1 & \textbackslash & 64 & \textbackslash & \textbackslash &\textbackslash \\
\hline
Number of channels 2 & \textbackslash & 128 & \textbackslash & \textbackslash & \textbackslash\\
\hline
Kernel size & \textbackslash & 3 & \textbackslash & \textbackslash & \textbackslash\\
\hline
\multirow{3}{*}{Pooling layer}
& \textbackslash & Pooling type: max-pooling & \textbackslash & \textbackslash & \textbackslash\\
& \textbackslash & Pool size:2 & \textbackslash & \textbackslash & \textbackslash\\
& \textbackslash & Stride:2 & \textbackslash & \textbackslash & \textbackslash\\
\bottomrule
\end{tabular}
}
\label{Table3}
\end{table*}
The experimental environment is Python 3.11, Pytorch 2.0.
The optimization function of all deep learning model training uses Adam, and the loss function uses MSE.
The dataset is divided into training set, validation set and test set in a ratio of 7:2:1.
The time step of input data is 12, and the prediction step is 1.
In order to ensure the effectiveness of the experiment, the parameters of the same model in serial and parallel structures remain unchanged.
Table \ref{Table3} shows the parameter settings of the models used in the experiment.
Table \ref{Table4} and Table \ref{Table5} are the comparison of the prediction performance of each baseline deep learning model in the validation set and test set data, respectively.
The results show that the prediction performance of CNN convolutional neural network is much better than that of BP network in the baseline model for information extraction of static features.
Among the baseline models for information extraction of dynamic features, GRU has the best prediction performance, followed by LSTM.
Therefore, CNN, GRU and LSTM are used to compare the performance of different structure fusion models.

\subsection{Comparison of power load forecasting performance of fusion model}\label{section: Comparison of power load forecasting performance of fusion model}
We will construct new fusion models that incorporate CNN, LSTM and GRU models in both serial and parallel configurations to extract static and dynamic features from power load data simultaneously.
In the serial mode, we input all the data into the model responsible for extracting static feature information, and the output is then fed into the model responsible for extracting dynamic feature information.
In the parallel mode, we simultaneously input the static part of the data into the model for extracting static information, and the dynamic part of the data into the model for extracting dynamic information.
Subsequently, we fuse the output results of the two models.
\setlength{\tabcolsep}{8pt}
\begin{table*}[htbp]\tiny
\caption{The prediction performance of each baseline deep learning model in the validation set.}
\centering
\scriptsize
\resizebox{0.6\textwidth}{!}{
\begin{tabular}{ccccc}
\toprule
Model &	MAPE & MAE & RMSE & R-Squared\\
\hline
BP & 1.866\% & 0.251 & 0.381 & 0.855\\
CNN & 0.988\% & 0.128 & 0.245 & 0.940\\
LSTM & 2.159\% & 0.316 & 0.401 & 0.836\\
BILSTM & 2.127\% & 0.315 & 0.400 & 0.839\\
GRU & 1.130\% & 0.180 & 0.259 & 0.933\\
\bottomrule
\end{tabular}
}
\label{Table4}
\end{table*}
\setlength{\tabcolsep}{8pt}
\begin{table*}[htbp]\tiny
\caption{The prediction performance of each baseline deep learning model in the test set.}
\centering
\scriptsize
\resizebox{0.6\textwidth}{!}{
\begin{tabular}{ccccc}
\toprule
Model & MAPE & MAE & RMSE & R-Squared\\
\hline
BP & 3.123\% & 0.124 & 0.225 & 0.829\\
CNN & 2.175\% & 0.099 & 0.143 & 0.930\\
LSTM & 3.457\% & 0.158 & 0.220 & 0.836\\
BILSTM & 3.947\% & 0.163 & 0.234 & 0.814\\
GRU & 2.613\% & 0.138 & 0.178 & 0.892\\
\bottomrule
\end{tabular}
}
\label{Table5}
\end{table*}
\setlength{\tabcolsep}{8pt}
\begin{table*}[htbp]
\caption{The prediction performance of each fusion model in the validation set.}
\centering
\resizebox{\textwidth}{!}{
\begin{tabular}{ccccc}
\toprule
Model & MAPE & MAE & RMSE & R-Squared\\
\hline
\multirow{2}{*}{SCL}
& 2.77\% & 0.344 & 0.450 & 0.798\\
& (+180.364\%, +28.300\%) & (+168.750\%, +8.861\%) & (+83.673\%, +12.219\%) & (-15.106\%, -4.545\%)\\
\multirow{2}{*}{PCL}
& 2.331\% & 0.329 & 0.417 & 0.826\\
& (+135.931\%, +7.967\%) & (+157.031\%, +4.114\%) & (+70.204\%, +3.990\%) & (-12.128\%, -1.196\%)\\
\multirow{2}{*}{SCG}
& 1.352\% & 0.167 & 0.248 & 0.938\\
& (+36.842\%, +19.646\%) & (+30.469\%, -7.222\%) & (+1.224\%, -4.247\%) & (-0.213\%, +0.536\%)\\
\multirow{2}{*}{PCG}
& 0.875\% & 0.133 & 0.220 & 0.952\\
& (-11.437\%, -22.566\%) & (+3.906\%, -26.111\%) & (-10.204\%, -15.058\%) & (+1.277\%, +2.036\%)\\
\multirow{2}{*}{SCGA}
& 0.931\% & 0.150 & 0.239 & 0.943\\
& (-5.769\%, -17.611\%) & (+17.187\%, -16.667\%) & (-2.449\%, -7.722\%) & (+0.319\%, +1.072\%)\\
\multirow{2}{*}{\textbf{PCGA}}
& \textbf{0.725\%} & \textbf{0.113} & \textbf{0.202} & \textbf{0.959}\\
& \textbf{(-26.194\%, -35.841\%)} & \textbf{(-11.719\%,-37.222\%)} & \textbf{(-17.551\%,-22.008\%)} & \textbf{(+2.021\%,+2.787\%)}\\
\bottomrule
\end{tabular}
}
\label{Table6}
\end{table*}
\setlength{\tabcolsep}{8pt}
\begin{table*}[htbp]
\caption{The prediction performance of each fusion model in the test set.}
\centering
\resizebox{\textwidth}{!}{
\begin{tabular}{ccccc}
\toprule
Model & MAPE & MAE & RMSE & R-Squared\\
\hline
\multirow{2}{*}{SCL}
& 4.199\% & 0.181 & 0.248 & 0.790\\
& (+93.057\%, +21.464\%) & (+82.828\%, +14.557\%) & (+73.427\%, +12.727\%) & (-15.054\%, -5.502\%)\\
\multirow{2}{*}{PCL}
& 4.058\% & 0.165 & 0.236 & 0.811\\
& (+86.575\%, +17.385\%) & (+66.667\%, +4.430\%) & (+65.035\%, +7.273\%) & (-12.796\%, -2.990\%)\\
\multirow{2}{*}{SCG}
& 2.216\% & 0.103 & 0.146 & 0.927\\
& (+1.885\%, -15.193\%) & (+4.040\%, -25.362\%) & (+2.098\%, -17.978\%) & (-0.323\%, +3.924\%)\\
\multirow{2}{*}{PCG}
& 1.509\% & 0.068 & 0.113 & 0.956\\
& (-30.621\%, -42.250\%) & (-31.313\%, -50.725\%) & (-20.979\%, -36.517\%) & (+2.796\%, +7.175\%)\\
\multirow{2}{*}{SCGA}
& 2.176\% & 0.100 & 0.142 & 0.931\\
& (+0.046\%, -16.724\%) & (+1.010\%, -27.536\%) & (-0.699\%, -20.225\%) & (+0.108\%, +4.372\%)\\
\multirow{2}{*}{\textbf{PCGA}}
& \textbf{1.453\%} & \textbf{0.081} & \textbf{0.104} & \textbf{0.963}\\
& \textbf{(-33.195\%, -44.393\%)} & \textbf{(-18.182\%, -41.304\%)} & \textbf{(-27.273\%,-41.573\%)} & \textbf{(+3.548\%, +7.960\%)}\\
\bottomrule
\end{tabular}
}
\label{Table7}
\end{table*}

Table \ref{Table6} and Table \ref{Table7} present the comparison of the prediction performance of each fusion model on the validation and test sets.
It can be seen that PCG achieves the highest prediction performance in the model without Attention layer, followed by SCG.
This can be attributed to the strong individual predictive capabilities of CNN and GRU model.
The prediction performance of PCG is better than that of SCG, and the prediction performance of PCL is better than that of SCL, demonstrating the strong information fusion capability of parallel structure.
We compare the prediction performance of the fusion model with the baseline model, and the change of each index is shown in parentheses.
Notably, the PCGA model exhibits the largest improvement in prediction performance on both the validation and test sets.
Figure \ref{fig7} provide the comparison of MAPE, MAE, RMSE and R-Squared indicators predicted by each model.
Among these indicators, the MAE value of the PCGA model on the test set is slightly higher than that of the PCG model, but the MAPE and RMSE values are lower compared to other models.
Additionally, the R-Squared value of the PCGA model is higher than other models.
Overall, the PCGA model demonstrates the best performance among all the models, followed by PCG model.

\begin{figure*}[ht]
  \centering
  \subfigure[]{
    \label{Fig8-a}
    \includegraphics[width=7.5cm]{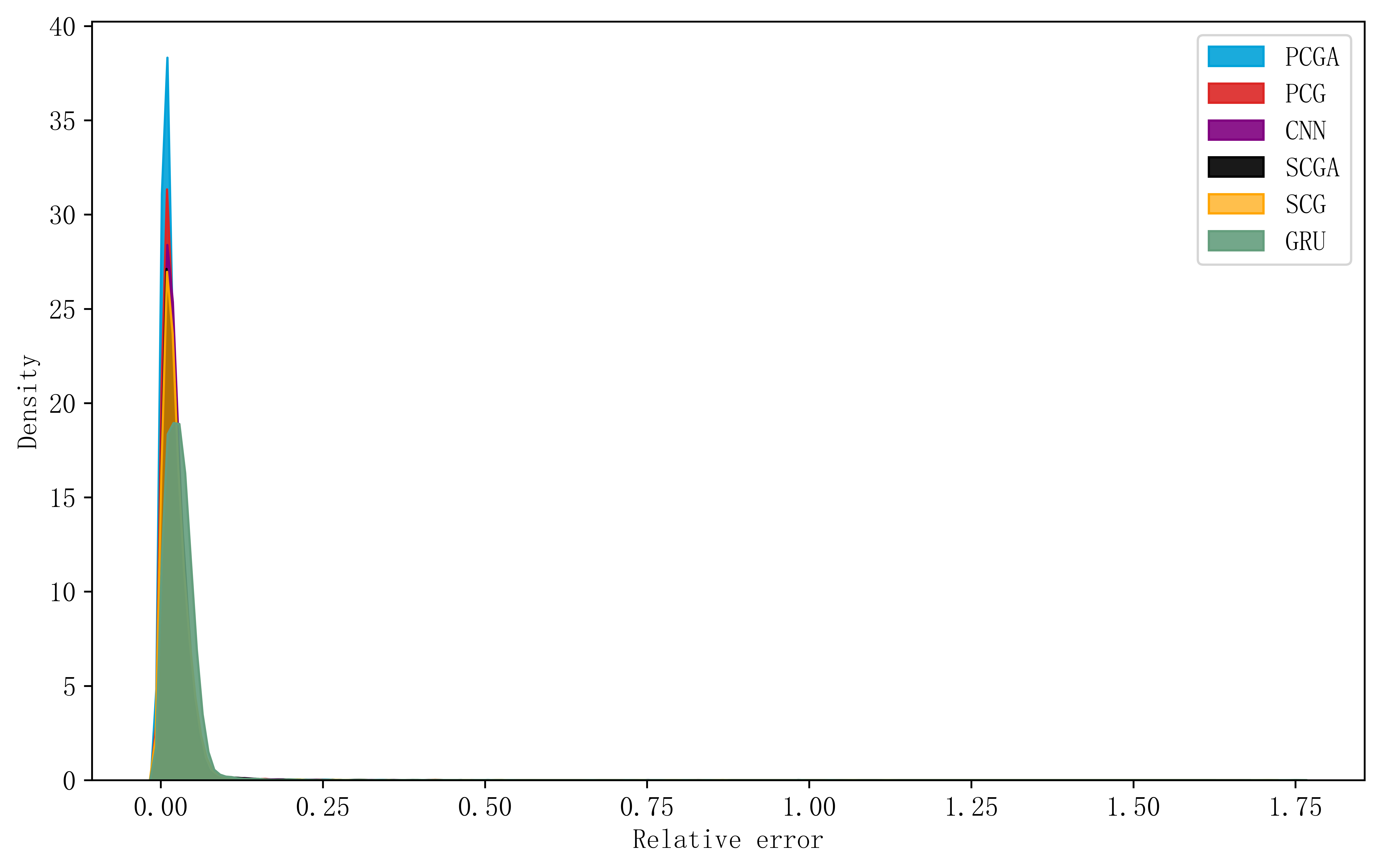}}
    \hspace{-0.1in}
  \subfigure[]{
    \label{Fig8-b}
    \includegraphics[width=8cm]{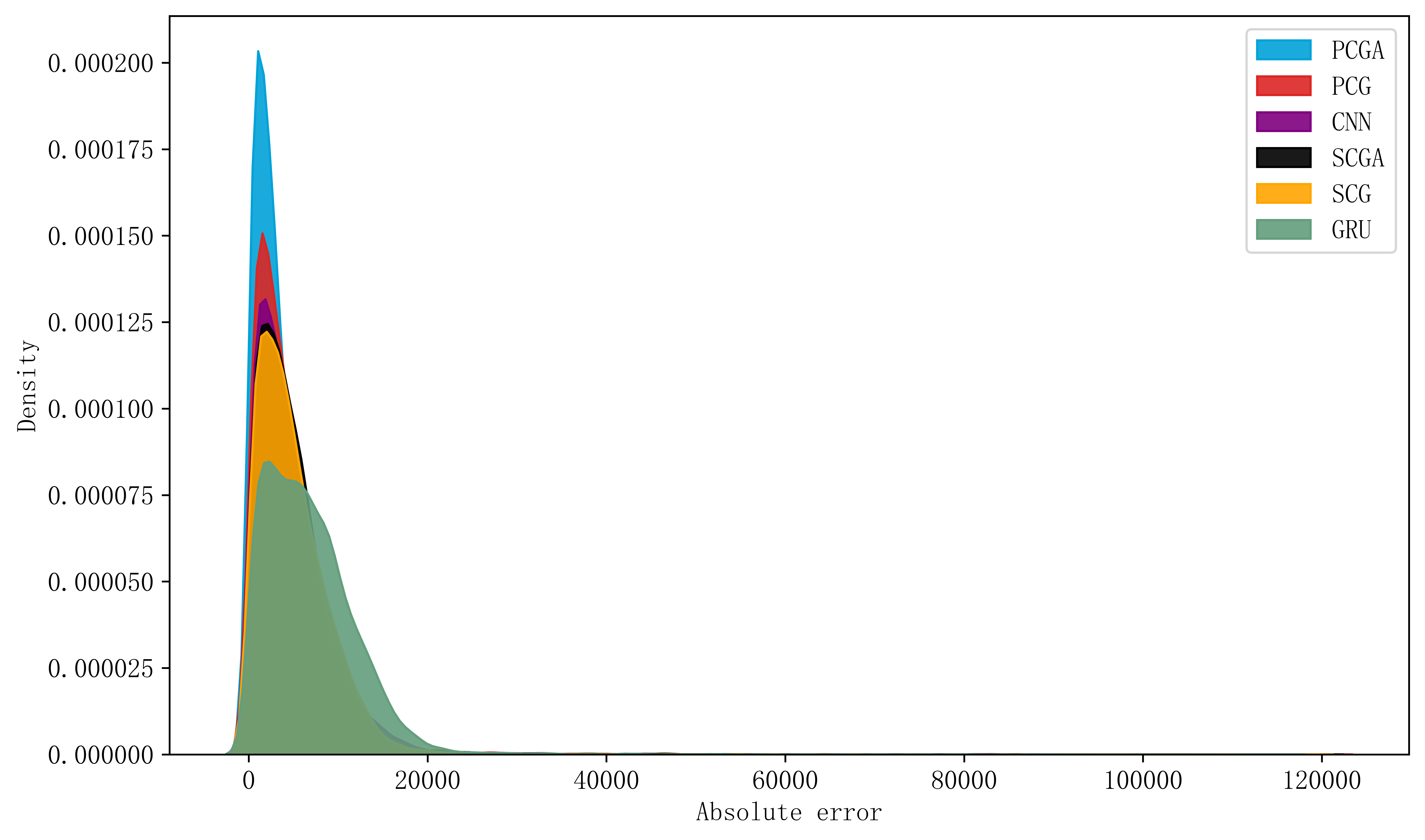}}
    \caption{The density map of the relative error (a) and absolute error (b) between the predicted results of each model and the actual load.}
  \label{fig8}
\end{figure*}
\begin{figure*}[htb]
  \centering
   \includegraphics[width=15.5cm]{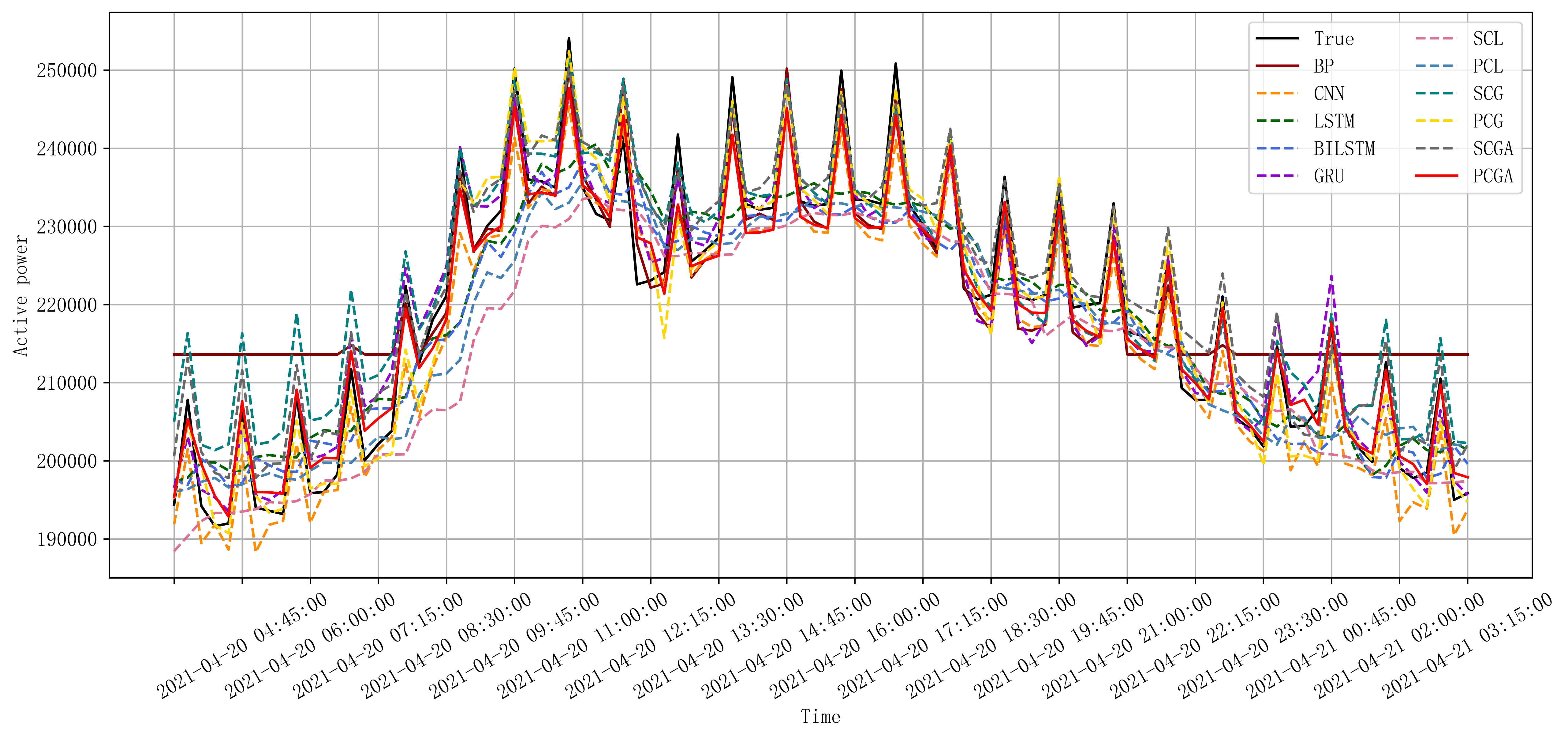}
   \caption{The prediction results of the most recent day of the test set of each model (from 3:45 on April 20, 2021 to 3:30 on April 21, 2021).}
    \label{fig10}
\end{figure*}
\begin{figure*}[htb]
  \centering
   \includegraphics[width=15.5cm]{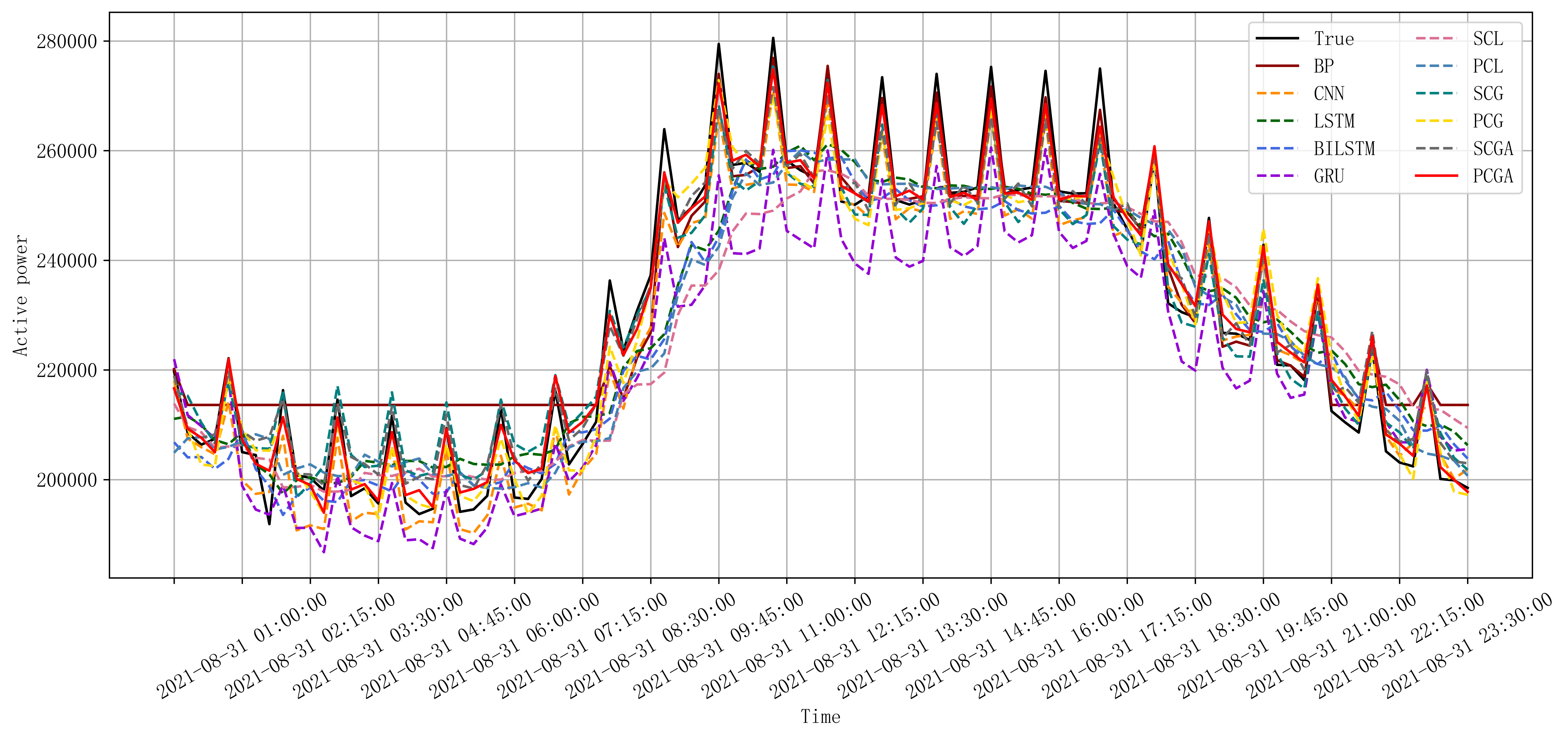}
   \caption{The prediction results of the farthest day of the test set of each model (from 3:45 on April 20, 2021 to 3:30 on April 21, 2021).}
    \label{fig11}
\end{figure*}
\subsection{Analysis of prediction results based on PCGA model}\label{section: Analysis of prediction results based on PCGA model}
We calculated the absolute and relative errors of the prediction results of each model on the test set and the original load, and use the density map to visually represent the comparison.
The density map can depict the probability density of data distribution and help to intuitively understand the error distribution, including the location of the error concentration, the degree of dispersion and the shape of the distribution.
Figure \ref{Fig8-a} shows the density map of relative errors between the predicted results of each model and the actual load, while figure \ref{Fig8-b} shows the density map of absolute errors between the predicted results of each model and the actual load.
It can be seen that both absolute error and relative error, the concentration of error values of the PCGA model is smaller than that of other models, and the dispersion degree of the error value is low, with no abnormal situations of large error.

In order to have more intuitive understanding of the error between the model prediction results and the actual data, we draw a line chart to directly display the predicted results of each model and actual data.
After the data set is divided in a ratio of 7:2:1, the test set data consists of 12,849 sample points from 3 :45 on April 20, 2021 to 23:45 on August 31, 2021.
Due to the large number of sample points, it is not clear to draw all of them.
We select the most recent (3:45 on April 20, 2021 to 3:30 on April 21, 2021) and the farthest (0:00 on August 31, 2021 to 23:45 on August 31, 2021) day, each with 96 sample points, and visualize the prediction results of each model, as shown in Figure \ref{fig10} and Figure \ref{fig11}, respectively.
Indeed, the local comparison reveals that the PCGA model provides predictions that are closest to the actual load values, particularly in terms of local maximum and minimum values.
The model exhibits smaller prediction deviations compared to other models, indicating its superior accuracy in capturing local fluctuations in the load data.
This suggests that the PCGA model is more effective in capturing the intricate patterns and variations in the data, leading to more accurate predictions in both peak and low load periods.

\section{Conclusion}\label{section: Conclusion}
Informed by the theory that the generalization ability of parallel structure models can be enhanced, this study presents a parallel CNN-GRU model with an optimized attention mechanism.
This model effectively enhances power consumption efficiency and ensures power system safety by accurately predicting power load.
The CNN captures the evolving characteristics of static spatial data, while the GRU captures the long-term dependencies in dynamic time series data.
Moreover, the attention mechanism is employed to identify significant features within the extracted spatio-temporal features.

Through experimental comparison, the proposed parallel CNN-LSTM model integrated with an attention mechanism, referred to as PCGA, demonstrates superior prediction accuracy compared to other models.
On the test set, the PCGA model achieves a mean absolute percentage error (MAPE) of 1.453\%, mean absolute error (MAE) of 0.081, root mean square error (RMSE) of 0.104, and R-squared (R2) value of 0.963.
Additionally, the PCGA model exhibits the smallest absolute error and relative error when comparing the predicted values to the actual load, outperforming all other models.

This paper introduces a novel research framework for power load forecasting.
Experimental results indicate that the fusion method employed in the model significantly impacts its performance.
It is observed that the parallel fusion method exhibits superior performance compared to the serial fusion method.
This can be attributed to the fact that the serial fusion method merely involves the straightforward input and output of the entire dataset, without considering the actual physical significance of the data or the applicability of the model.
In contrast, the parallel fusion method first divides the input data into static and dynamic components, and subsequently employs suitable models to extract their crucial temporal and spatial features.
These features are then fused, aligning with the actual underlying mechanism.
Furthermore, we proved theoretically that the parallel fusion method can improve the performance of the model more than the serial fusion method.

\section*{CRediT authorship contribution statement}
Chao Min: Conceptualization, Methodology, Supervision, Writing -- review. Yijia Wang: Data curation, Software, Writing -- original draft \& editing. Bo Zhang: Writing –review \& editing. Xin Ma: Writing –review \& editing. Junyi Cui: Writing –review \& editing.

\section*{Acknowledgments}

\printcredits

\bibliographystyle{cas-model2-names}

\bibliography{My_EndNote_Library_N}

\begin{thebibliography}{53}
\expandafter\ifx\csname natexlab\endcsname\relax\def\natexlab#1{#1}\fi
\providecommand{\url}[1]{\texttt{#1}}
\providecommand{\href}[2]{#2}
\providecommand{\path}[1]{#1}
\providecommand{\DOIprefix}{doi:}
\providecommand{\ArXivprefix}{arXiv:}
\providecommand{\URLprefix}{URL: }
\providecommand{\Pubmedprefix}{pmid:}
\providecommand{\doi}[1]{\href{http://dx.doi.org/#1}{\path{#1}}}
\providecommand{\Pubmed}[1]{\href{pmid:#1}{\path{#1}}}
\providecommand{\bibinfo}[2]{#2}
\ifx\xfnm\relax \def\xfnm[#1]{\unskip,\space#1}\fi
\bibitem[{Alghamdi et~al.(2023)Alghamdi, Hafeez, Ali, Ullah, Khan, Murawwat and
  Hua}]{Alghamdi2023AnIM}
\bibinfo{author}{Alghamdi, H.}, \bibinfo{author}{Hafeez, G.},
  \bibinfo{author}{Ali, S.}, \bibinfo{author}{Ullah, S.},
  \bibinfo{author}{Khan, M.I.}, \bibinfo{author}{Murawwat, S.},
  \bibinfo{author}{Hua, L.G.}, \bibinfo{year}{2023}.
\newblock \bibinfo{title}{An integrated model of deep learning and heuristic
  algorithm for load forecasting in smart grid}.
\newblock \bibinfo{journal}{Mathematics} \URLprefix
  \url{https://api.semanticscholar.org/CorpusID:265057891}.
\bibitem[{Arrieta et~al.(2019)Arrieta, Rodr{\'i}guez, Ser, Bennetot, Tabik,
  Barbado, Garc{\'i}a, Gil-Lopez, Molina, Benjamins, Chatila and
  Herrera}]{Arrieta2019ExplainableAI}
\bibinfo{author}{Arrieta, A.B.}, \bibinfo{author}{Rodr{\'i}guez, N.D.},
  \bibinfo{author}{Ser, J.D.}, \bibinfo{author}{Bennetot, A.},
  \bibinfo{author}{Tabik, S.}, \bibinfo{author}{Barbado, A.},
  \bibinfo{author}{Garc{\'i}a, S.}, \bibinfo{author}{Gil-Lopez, S.},
  \bibinfo{author}{Molina, D.}, \bibinfo{author}{Benjamins, R.},
  \bibinfo{author}{Chatila, R.}, \bibinfo{author}{Herrera, F.},
  \bibinfo{year}{2019}.
\newblock \bibinfo{title}{Explainable artificial intelligence (xai): Concepts,
  taxonomies, opportunities and challenges toward responsible ai}.
\newblock \bibinfo{journal}{Inf. Fusion} \bibinfo{volume}{58},
  \bibinfo{pages}{82--115}.
\newblock \URLprefix \url{https://api.semanticscholar.org/CorpusID:204824113}.
\bibitem[{Bartlett and Mendelson(2003)}]{Bartlett2003RademacherAG}
\bibinfo{author}{Bartlett, P.L.}, \bibinfo{author}{Mendelson, S.},
  \bibinfo{year}{2003}.
\newblock \bibinfo{title}{Rademacher and gaussian complexities: Risk bounds and
  structural results}, in: \bibinfo{booktitle}{Journal of machine learning
  research}.
\newblock \URLprefix \url{https://api.semanticscholar.org/CorpusID:463216}.
\bibitem[{Belhaiza and Al-Abdallah(2024)}]{Belhaiza2024NeuralNetwork}
\bibinfo{author}{Belhaiza, S.}, \bibinfo{author}{Al-Abdallah, S.},
  \bibinfo{year}{2024}.
\newblock \bibinfo{title}{A neural network forecasting approach for the smart
  grid demand response management problem}.
\newblock \bibinfo{journal}{Energies} \bibinfo{volume}{17},
  \bibinfo{pages}{2329}.
\bibitem[{Bizrah and Al-Muhaini(2017)}]{Bizrah2017TheIO}
\bibinfo{author}{Bizrah, A.}, \bibinfo{author}{Al-Muhaini, M.},
  \bibinfo{year}{2017}.
\newblock \bibinfo{title}{The impact of seasonal arma wind speed modeling on
  the reliability of power distribution systems}.
\newblock \bibinfo{journal}{2017 IEEE Power $\&$ Energy Society General
  Meeting} , \bibinfo{pages}{1--5}\URLprefix
  \url{https://api.semanticscholar.org/CorpusID:34288908}.
\bibitem[{Běl{\'i}k and Rubanenko(2023)}]{Blk2023ImplementationOD}
\bibinfo{author}{Běl{\'i}k, M.}, \bibinfo{author}{Rubanenko, O.},
  \bibinfo{year}{2023}.
\newblock \bibinfo{title}{Implementation of digital twin for increasing
  efficiency of renewable energy sources}.
\newblock \bibinfo{journal}{Energies} \URLprefix
  \url{https://api.semanticscholar.org/CorpusID:259430317}.
\bibitem[{he~Chen et~al.(2023)he~Chen, Zhu, Hu, Wang, Sun, Yang, Li and
  Meng}]{Chen2023MultifeatureSP}
\bibinfo{author}{he~Chen, H.}, \bibinfo{author}{Zhu, M.}, \bibinfo{author}{Hu,
  X.}, \bibinfo{author}{Wang, J.}, \bibinfo{author}{Sun, Y.},
  \bibinfo{author}{Yang, J.}, \bibinfo{author}{Li, B.}, \bibinfo{author}{Meng,
  X.}, \bibinfo{year}{2023}.
\newblock \bibinfo{title}{Multifeature short-term power load forecasting based
  on gcn-lstm}.
\newblock \bibinfo{journal}{International Transactions on Electrical Energy
  Systems} \URLprefix \url{https://api.semanticscholar.org/CorpusID:261378595}.
\bibitem[{Chen et~al.(2021)Chen, Feng, Jiang and Zhu}]{Chen2021StateOC}
\bibinfo{author}{Chen, J.}, \bibinfo{author}{Feng, X.}, \bibinfo{author}{Jiang,
  L.}, \bibinfo{author}{Zhu, Q.}, \bibinfo{year}{2021}.
\newblock \bibinfo{title}{State of charge estimation of lithium-ion battery
  using denoising autoencoder and gated recurrent unit recurrent neural
  network}.
\newblock \bibinfo{journal}{Energy} \URLprefix
  \url{https://api.semanticscholar.org/CorpusID:233682520}.
\bibitem[{Chen et~al.(2024)Chen, Lin, Zhang, Liu and Yu}]{Chen2024DayaheadLF}
\bibinfo{author}{Chen, Y.}, \bibinfo{author}{Lin, C.}, \bibinfo{author}{Zhang,
  Y.}, \bibinfo{author}{Liu, J.}, \bibinfo{author}{Yu, D.},
  \bibinfo{year}{2024}.
\newblock \bibinfo{title}{Day-ahead load forecast based on conv2d-gru\_sc aimed
  to adapt to steep changes in load}.
\newblock \bibinfo{journal}{Energy} \URLprefix
  \url{https://api.semanticscholar.org/CorpusID:270058095}.
\bibitem[{Chernoff and Herman(1952)}]{Chernoff1952A}
\bibinfo{author}{Chernoff}, \bibinfo{author}{Herman}, \bibinfo{year}{1952}.
\newblock \bibinfo{title}{A measure of asymptotic efficiency for tests of a
  hypothesis based on the sum of observations}.
\newblock \bibinfo{journal}{Annals of Mathematical Statistics}
  \bibinfo{volume}{23}, \bibinfo{pages}{493--507}.
\bibitem[{Chopra et~al.(2018)Chopra, Murthy and
  Rangarajan}]{Chopra2018StatisticalTF}
\bibinfo{author}{Chopra, R.}, \bibinfo{author}{Murthy, C.R.},
  \bibinfo{author}{Rangarajan, G.}, \bibinfo{year}{2018}.
\newblock \bibinfo{title}{Statistical tests for detecting granger causality}.
\newblock \bibinfo{journal}{IEEE Transactions on Signal Processing}
  \bibinfo{volume}{66}, \bibinfo{pages}{5803--5816}.
\newblock \URLprefix \url{https://api.semanticscholar.org/CorpusID:52840312}.
\bibitem[{Dang et~al.(2022)Dang, Peng, Zhao, Li and Kong}]{Dang2022AQR}
\bibinfo{author}{Dang, S.}, \bibinfo{author}{Peng, L.}, \bibinfo{author}{Zhao,
  J.}, \bibinfo{author}{Li, J.}, \bibinfo{author}{Kong, Z.},
  \bibinfo{year}{2022}.
\newblock \bibinfo{title}{A quantile regression random forest-based short-term
  load probabilistic forecasting method}.
\newblock \bibinfo{journal}{Energies} \URLprefix
  \url{https://api.semanticscholar.org/CorpusID:250603073}.
\bibitem[{Deebani and Nezamoddini-Kachouie(2020)}]{Deebani2020MonteCE}
\bibinfo{author}{Deebani, W.}, \bibinfo{author}{Nezamoddini-Kachouie, N.},
  \bibinfo{year}{2020}.
\newblock \bibinfo{title}{Monte carlo ensemble correlation coefficient for
  association detection}.
\newblock \bibinfo{journal}{Communications in Statistics - Simulation and
  Computation} \bibinfo{volume}{51}, \bibinfo{pages}{7095 -- 7109}.
\newblock \URLprefix \url{https://api.semanticscholar.org/CorpusID:224844677}.
\bibitem[{Gao et~al.(2022)Gao, Niu, Ji and Sun}]{Gao2022MidtermED}
\bibinfo{author}{Gao, T.}, \bibinfo{author}{Niu, D.}, \bibinfo{author}{Ji, Z.},
  \bibinfo{author}{Sun, L.}, \bibinfo{year}{2022}.
\newblock \bibinfo{title}{Mid-term electricity demand forecasting using
  improved variational mode decomposition and extreme learning machine
  optimized by sparrow search algorithm}.
\newblock \bibinfo{journal}{Energy} \URLprefix
  \url{https://api.semanticscholar.org/CorpusID:252075268}.
\bibitem[{Huang and Qin(2022)}]{Huang2022InfluenceFC}
\bibinfo{author}{Huang, Z.}, \bibinfo{author}{Qin, G.}, \bibinfo{year}{2022}.
\newblock \bibinfo{title}{Influence function-based confidence intervals for the
  kendall rank correlation coefficient}.
\newblock \bibinfo{journal}{Computational Statistics} \bibinfo{volume}{38},
  \bibinfo{pages}{1041--1055}.
\newblock \URLprefix \url{https://api.semanticscholar.org/CorpusID:251353871}.
\bibitem[{Ibrahim et~al.(2022)Ibrahim, Rabelo, Gutierrez-Franco and
  Clavijo-Buritica}]{Ibrahim2022MachineLF}
\bibinfo{author}{Ibrahim, B.}, \bibinfo{author}{Rabelo, L.C.},
  \bibinfo{author}{Gutierrez-Franco, E.}, \bibinfo{author}{Clavijo-Buritica,
  N.}, \bibinfo{year}{2022}.
\newblock \bibinfo{title}{Machine learning for short-term load forecasting in
  smart grids}.
\newblock \bibinfo{journal}{Energies} \URLprefix
  \url{https://api.semanticscholar.org/CorpusID:253288812}.
\bibitem[{Jafari et~al.(2023)Jafari, Lai and Yanushkevich}]{Jafari2023MVARAC}
\bibinfo{author}{Jafari, B.}, \bibinfo{author}{Lai, K.},
  \bibinfo{author}{Yanushkevich, S.N.}, \bibinfo{year}{2023}.
\newblock \bibinfo{title}{Mvar and causal modeling of relationship between
  physiological signals and affective states}.
\newblock \bibinfo{journal}{2023 IEEE Conference on Artificial Intelligence
  (CAI)} , \bibinfo{pages}{134--135}\URLprefix
  \url{https://api.semanticscholar.org/CorpusID:260387952}.
\bibitem[{Jiang et~al.(2021)Jiang, Huang, Zhang, Lin, Zhang, Hu, Liu, Jiang,
  Yang and Lin}]{Jiang2021MediumlongTL}
\bibinfo{author}{Jiang, Y.}, \bibinfo{author}{Huang, Q.},
  \bibinfo{author}{Zhang, K.}, \bibinfo{author}{Lin, Z.},
  \bibinfo{author}{Zhang, T.}, \bibinfo{author}{Hu, X.}, \bibinfo{author}{Liu,
  S.}, \bibinfo{author}{Jiang, C.X.}, \bibinfo{author}{Yang, L.},
  \bibinfo{author}{Lin, Z.}, \bibinfo{year}{2021}.
\newblock \bibinfo{title}{Medium-long term load forecasting method considering
  industry correlation for power management}.
\newblock \bibinfo{journal}{Energy Reports} \URLprefix
  \url{https://api.semanticscholar.org/CorpusID:244940281}.
\bibitem[{Karunasingha(2021)}]{Karunasingha2021RootMS}
\bibinfo{author}{Karunasingha, D.S.K.}, \bibinfo{year}{2021}.
\newblock \bibinfo{title}{Root mean square error or mean absolute error? use
  their ratio as well}.
\newblock \bibinfo{journal}{Inf. Sci.} \bibinfo{volume}{585},
  \bibinfo{pages}{609--629}.
\newblock \URLprefix \url{https://api.semanticscholar.org/CorpusID:244741650}.
\bibitem[{Kim et~al.(2023)Kim, Yamaguchi and Shimoda}]{Kim2023PhysicsbasedMO}
\bibinfo{author}{Kim, B.}, \bibinfo{author}{Yamaguchi, Y.},
  \bibinfo{author}{Shimoda, Y.}, \bibinfo{year}{2023}.
\newblock \bibinfo{title}{Physics-based modeling of electricity load profile of
  commercial building stock considering building system composition and
  occupancy profile}.
\newblock \bibinfo{journal}{Energy and Buildings} .
\bibitem[{Kim and Cho(2019)}]{Kim2019PredictingRE}
\bibinfo{author}{Kim, T.Y.}, \bibinfo{author}{Cho, S.B.}, \bibinfo{year}{2019}.
\newblock \bibinfo{title}{Predicting residential energy consumption using
  cnn-lstm neural networks}.
\newblock \bibinfo{journal}{Energy} \URLprefix
  \url{https://api.semanticscholar.org/CorpusID:195394929}.
\bibitem[{Kolster et~al.(2022)Kolster, Niessen and Duckheim}]{2022Providing}
\bibinfo{author}{Kolster, T.}, \bibinfo{author}{Niessen, S.},
  \bibinfo{author}{Duckheim, M.}, \bibinfo{year}{2022}.
\newblock \bibinfo{title}{Providing distributed flexibility for curative
  transmission system operation using a scalable robust optimization approach}.
\newblock \bibinfo{journal}{Electric Power Systems Research} .
\bibitem[{Li et~al.(2009)Li, Fu, Li and Zhang}]{Li2009TheIT}
\bibinfo{author}{Li, Y.}, \bibinfo{author}{Fu, Y.}, \bibinfo{author}{Li, H.},
  \bibinfo{author}{Zhang, S.}, \bibinfo{year}{2009}.
\newblock \bibinfo{title}{The improved training algorithm of back propagation
  neural network with self-adaptive learning rate}.
\newblock \bibinfo{journal}{2009 International Conference on Computational
  Intelligence and Natural Computing} \bibinfo{volume}{1},
  \bibinfo{pages}{73--76}.
\newblock \URLprefix \url{https://api.semanticscholar.org/CorpusID:10557754}.
\bibitem[{Liu et~al.(2021)Liu, Xiao, Li, Wang, Bie and
  Jiao}]{Liu2021SimulationOD}
\bibinfo{author}{Liu, J.}, \bibinfo{author}{Xiao, B.}, \bibinfo{author}{Li,
  Y.}, \bibinfo{author}{Wang, X.}, \bibinfo{author}{Bie, Q.},
  \bibinfo{author}{Jiao, J.}, \bibinfo{year}{2021}.
\newblock \bibinfo{title}{Simulation of dynamic urban expansion under
  ecological constraints using a long short term memory network model and
  cellular automata}.
\newblock \bibinfo{journal}{Remote. Sens.} \bibinfo{volume}{13},
  \bibinfo{pages}{1499}.
\newblock \URLprefix \url{https://api.semanticscholar.org/CorpusID:234823453}.
\bibitem[{Luo(2017)}]{Luo2017ShortTP}
\bibinfo{author}{Luo, J.}, \bibinfo{year}{2017}.
\newblock \bibinfo{title}{Short term power load forecasting considering
  meteorological factors}.
\newblock \URLprefix \url{https://api.semanticscholar.org/CorpusID:184914344}.
\bibitem[{Lv et~al.(2018)Lv, Cheng, YanShuang and Tang}]{Lv2018Shortterm}
\bibinfo{author}{Lv, X.}, \bibinfo{author}{Cheng, X.},
  \bibinfo{author}{YanShuang}, \bibinfo{author}{Tang, Y.m.},
  \bibinfo{year}{2018}.
\newblock \bibinfo{title}{Short-term power load forecasting based on balanced
  knn}.
\newblock \bibinfo{journal}{IOP Conference Series: Materials Science and
  Engineering} \bibinfo{volume}{322}, \bibinfo{pages}{072058}.
\newblock \DOIprefix\doi{10.1088/1757-899X/322/7/072058}.
\bibitem[{Madhukumar et~al.(2022)Madhukumar, Sebastian, Liang, Jamil and
  Shabbir}]{Madhukumar2022RegressionMS}
\bibinfo{author}{Madhukumar, M.}, \bibinfo{author}{Sebastian, A.},
  \bibinfo{author}{Liang, X.}, \bibinfo{author}{Jamil, M.},
  \bibinfo{author}{Shabbir, M.N.S.K.}, \bibinfo{year}{2022}.
\newblock \bibinfo{title}{Regression model-based short-term load forecasting
  for university campus load}.
\newblock \bibinfo{journal}{IEEE Access} \bibinfo{volume}{10},
  \bibinfo{pages}{8891--8905}.
\newblock \URLprefix \url{https://api.semanticscholar.org/CorpusID:246050177}.
\bibitem[{Mitchell(2003)}]{2003Machine}
\bibinfo{author}{Mitchell, T.}, \bibinfo{year}{2003}.
\newblock \bibinfo{title}{Machine Learning}.
\newblock \bibinfo{publisher}{Machine Learning}.
\bibitem[{Mortezanejad et~al.(2019)Mortezanejad, Borzadaran and
  Gildeh}]{Mortezanejad2019JointDD}
\bibinfo{author}{Mortezanejad, S.A.F.}, \bibinfo{author}{Borzadaran, G.R.M.},
  \bibinfo{author}{Gildeh, B.S.}, \bibinfo{year}{2019}.
\newblock \bibinfo{title}{Joint dependence distribution of data set using
  optimizing tsallis copula entropy}.
\newblock \bibinfo{journal}{Physica A: Statistical Mechanics and its
  Applications} \URLprefix
  \url{https://api.semanticscholar.org/CorpusID:198443920}.
\bibitem[{Myttenaere et~al.(2016)Myttenaere, Golden, Grand and
  Rossi}]{Myttenaere2016MeanAP}
\bibinfo{author}{Myttenaere, A.D.}, \bibinfo{author}{Golden, B.},
  \bibinfo{author}{Grand, B.L.}, \bibinfo{author}{Rossi, F.},
  \bibinfo{year}{2016}.
\newblock \bibinfo{title}{Mean absolute percentage error for regression
  models}.
\newblock \bibinfo{journal}{Neurocomputing} \bibinfo{volume}{192},
  \bibinfo{pages}{38--48}.
\newblock \URLprefix \url{https://api.semanticscholar.org/CorpusID:44496422}.
\bibitem[{Nazeer et~al.(2019)Nazeer, Javaid, Khan, Hussain, Basheer and
  Ratyal}]{Nazeer2019ShortTerm}
\bibinfo{author}{Nazeer, O.}, \bibinfo{author}{Javaid, N.},
  \bibinfo{author}{Khan, A.B.M.}, \bibinfo{author}{Hussain, A.},
  \bibinfo{author}{Basheer, T.}, \bibinfo{author}{Ratyal, M.M.A.},
  \bibinfo{year}{2019}.
\newblock \bibinfo{title}{Short term load forcasting using heuristic algorithm
  and support vector machine}, in: \bibinfo{booktitle}{Complex, Intelligent,
  and Software Intensive Systems}, pp. \bibinfo{pages}{791--799}.
\bibitem[{Omaji et~al.(2020)Omaji, Javaid, Khalid, Khan and Kim}]{2020Towards}
\bibinfo{author}{Omaji, S.}, \bibinfo{author}{Javaid, N.},
  \bibinfo{author}{Khalid, A.}, \bibinfo{author}{Khan, W.Z.},
  \bibinfo{author}{Kim, B.S.}, \bibinfo{year}{2020}.
\newblock \bibinfo{title}{Towards real-time energy management of
  multi-microgrid using a deep convolution neural network and cooperative game
  approach}.
\newblock \bibinfo{journal}{IEEE Access} .
\bibitem[{Pansota et~al.(2021)Pansota, Javed, Muqeet, Irfan, Shehzad and
  Liaqat}]{Pansota2021SchedulingAS}
\bibinfo{author}{Pansota, M.S.}, \bibinfo{author}{Javed, H.},
  \bibinfo{author}{Muqeet, A.}, \bibinfo{author}{Irfan, M.},
  \bibinfo{author}{Shehzad, M.}, \bibinfo{author}{Liaqat, R.},
  \bibinfo{year}{2021}.
\newblock \bibinfo{title}{Scheduling and sizing of campus microgrid considering
  demand response and economic analysis}.
\newblock \bibinfo{journal}{Sensors (Basel, Switzerland)} \bibinfo{volume}{22}.
\newblock \URLprefix \url{https://api.semanticscholar.org/CorpusID:239434569}.
\bibitem[{Qi et~al.(2020)Qi, Du, Siniscalchi, Ma and Lee}]{Qi2020OnMA}
\bibinfo{author}{Qi, J.}, \bibinfo{author}{Du, J.},
  \bibinfo{author}{Siniscalchi, S.M.}, \bibinfo{author}{Ma, X.},
  \bibinfo{author}{Lee, C.H.}, \bibinfo{year}{2020}.
\newblock \bibinfo{title}{On mean absolute error for deep neural network based
  vector-to-vector regression}.
\newblock \bibinfo{journal}{IEEE Signal Processing Letters}
  \bibinfo{volume}{27}, \bibinfo{pages}{1485--1489}.
\newblock \URLprefix \url{https://api.semanticscholar.org/CorpusID:221139276}.
\bibitem[{Qi et~al.(2017)Qi, Luo, Wang and Wu}]{Qi2017LoadPR}
\bibinfo{author}{Qi, Y.}, \bibinfo{author}{Luo, B.}, \bibinfo{author}{Wang,
  X.}, \bibinfo{author}{Wu, L.}, \bibinfo{year}{2017}.
\newblock \bibinfo{title}{Load pattern recognition method based on fuzzy
  clustering and decision tree}.
\newblock \bibinfo{journal}{2017 IEEE Conference on Energy Internet and Energy
  System Integration (EI2)} , \bibinfo{pages}{1--5}\URLprefix
  \url{https://api.semanticscholar.org/CorpusID:6956071}.
\bibitem[{Sadaei et~al.(2019)Sadaei, de~Lima~e Silva, Guimar{\~a}es and
  Lee}]{Sadaei2019ShorttermLF}
\bibinfo{author}{Sadaei, H.J.}, \bibinfo{author}{de~Lima~e Silva, P.C.},
  \bibinfo{author}{Guimar{\~a}es, F.G.}, \bibinfo{author}{Lee, M.H.},
  \bibinfo{year}{2019}.
\newblock \bibinfo{title}{Short-term load forecasting by using a combined
  method of convolutional neural networks and fuzzy time series}.
\newblock \bibinfo{journal}{Energy} \URLprefix
  \url{https://api.semanticscholar.org/CorpusID:116610103}.
\bibitem[{Sajjad et~al.(2020)Sajjad, Khan, Ullah, Hussain, Ullah, Lee and
  Baik}]{Sajjad2020ANC}
\bibinfo{author}{Sajjad, M.}, \bibinfo{author}{Khan, Z.A.},
  \bibinfo{author}{Ullah, A.}, \bibinfo{author}{Hussain, T.},
  \bibinfo{author}{Ullah, W.}, \bibinfo{author}{Lee, M.Y.},
  \bibinfo{author}{Baik, S.W.}, \bibinfo{year}{2020}.
\newblock \bibinfo{title}{A novel cnn-gru-based hybrid approach for short-term
  residential load forecasting}.
\newblock \bibinfo{journal}{IEEE Access} \bibinfo{volume}{8},
  \bibinfo{pages}{143759--143768}.
\newblock \URLprefix \url{https://api.semanticscholar.org/CorpusID:221161846}.
\bibitem[{Shuping et~al.(2022)Shuping, Zhongming, Jing, Dahai, Yan and
  Ziyue}]{2022Fisher}
\bibinfo{author}{Shuping, C.}, \bibinfo{author}{Zhongming, S.},
  \bibinfo{author}{Jing, Y.}, \bibinfo{author}{Dahai, T.},
  \bibinfo{author}{Yan, C.}, \bibinfo{author}{Ziyue, Z.}, \bibinfo{year}{2022}.
\newblock \bibinfo{title}{Fisher information and online svr-based dynamic
  modeling methodology for meteorological sensitive load forecasting in smart
  grids}.
\newblock \bibinfo{journal}{Electrical engineering} , \bibinfo{pages}{104}.
\bibitem[{Singhal et~al.(2019)Singhal, Choudhary and
  Singh}]{Singhal2019ShortTermLF}
\bibinfo{author}{Singhal, R.}, \bibinfo{author}{Choudhary, N.K.},
  \bibinfo{author}{Singh, N.}, \bibinfo{year}{2019}.
\newblock \bibinfo{title}{Short-term load forecasting using hybrid arima and
  artificial neural network model}.
\newblock \URLprefix \url{https://api.semanticscholar.org/CorpusID:214578867}.
\bibitem[{Smyl et~al.(2022)Smyl, Dudek and Pełka}]{Smyl2022ESdRNNWD}
\bibinfo{author}{Smyl, S.}, \bibinfo{author}{Dudek, G.},
  \bibinfo{author}{Pełka, P.}, \bibinfo{year}{2022}.
\newblock \bibinfo{title}{Es-drnn with dynamic attention for short-term load
  forecasting}.
\newblock \bibinfo{journal}{2022 International Joint Conference on Neural
  Networks (IJCNN)} , \bibinfo{pages}{1--8}\URLprefix
  \url{https://api.semanticscholar.org/CorpusID:247218448}.
\bibitem[{Song et~al.(2024)Song, Chen, Wang, Zhang and Sun}]{Song2024AML}
\bibinfo{author}{Song, X.}, \bibinfo{author}{Chen, Z.}, \bibinfo{author}{Wang,
  J.}, \bibinfo{author}{Zhang, Y.}, \bibinfo{author}{Sun, X.},
  \bibinfo{year}{2024}.
\newblock \bibinfo{title}{A multi-stage lstm federated forecasting method for
  multi-loads under multi-time scales}.
\newblock \bibinfo{journal}{Expert Systems with Applications} \URLprefix
  \url{https://api.semanticscholar.org/CorpusID:270164322}.
\bibitem[{Sun et~al.(2021)Sun, Qin and
  Zhuang}]{Sun2021NonparametriccopulaentropyAN}
\bibinfo{author}{Sun, Y.}, \bibinfo{author}{Qin, W.}, \bibinfo{author}{Zhuang,
  Z.}, \bibinfo{year}{2021}.
\newblock \bibinfo{title}{Nonparametric-copula-entropy and network
  deconvolution method for causal discovery in complex manufacturing systems}.
\newblock \bibinfo{journal}{Journal of Intelligent Manufacturing}
  \bibinfo{volume}{33}, \bibinfo{pages}{1699 -- 1713}.
\newblock \URLprefix \url{https://api.semanticscholar.org/CorpusID:233674438}.
\bibitem[{Tang and Cai(2023)}]{Tang2023ShortTermPL}
\bibinfo{author}{Tang, Y.}, \bibinfo{author}{Cai, H.}, \bibinfo{year}{2023}.
\newblock \bibinfo{title}{Short-term power load forecasting based on
  vmd-pyraformer-adan}.
\newblock \bibinfo{journal}{IEEE Access} \bibinfo{volume}{11},
  \bibinfo{pages}{61958--61967}.
\newblock \URLprefix \url{https://api.semanticscholar.org/CorpusID:258585708}.
\bibitem[{Tian et~al.(2022)Tian, Ye, Lou, Zuo, Zhang and
  Li}]{Tian2022Dailypower}
\bibinfo{author}{Tian, C.}, \bibinfo{author}{Ye, Y.}, \bibinfo{author}{Lou,
  Y.}, \bibinfo{author}{Zuo, W.}, \bibinfo{author}{Zhang, G.},
  \bibinfo{author}{Li, C.}, \bibinfo{year}{2022}.
\newblock \bibinfo{title}{Daily power demand prediction for buildings at a
  large scale using a hybrid of physics-based model and generative adversarial
  network}.
\newblock \bibinfo{journal}{Building Simulation} , \bibinfo{pages}{1685--1701}.
\bibitem[{Wang(2016)}]{Wang2016ImprovedST}
\bibinfo{author}{Wang, W.}, \bibinfo{year}{2016}.
\newblock \bibinfo{title}{Improved short term load forecasting of power system
  based on arma model}.
\newblock \URLprefix \url{https://api.semanticscholar.org/CorpusID:57170537}.
\bibitem[{Wenlong and Yahui(2020)}]{Wenlong2020LoadFO}
\bibinfo{author}{Wenlong, H.}, \bibinfo{author}{Yahui, W.},
  \bibinfo{year}{2020}.
\newblock \bibinfo{title}{Load forecast of gas region based on arima
  algorithm}.
\newblock \bibinfo{journal}{2020 Chinese Control And Decision Conference
  (CCDC)} , \bibinfo{pages}{1960--1965}\URLprefix
  \url{https://api.semanticscholar.org/CorpusID:221120857}.
\bibitem[{Wu et~al.(2020)Wu, Jiang, Zhou, Wang, Zuo, Wu, Liang and
  Liu}]{Wu2020ApplicationOI}
\bibinfo{author}{Wu, Z.}, \bibinfo{author}{Jiang, S.}, \bibinfo{author}{Zhou,
  X.}, \bibinfo{author}{Wang, Y.}, \bibinfo{author}{Zuo, Y.},
  \bibinfo{author}{Wu, Z.}, \bibinfo{author}{Liang, L.}, \bibinfo{author}{Liu,
  Q.}, \bibinfo{year}{2020}.
\newblock \bibinfo{title}{Application of image retrieval based on convolutional
  neural networks and hu invariant moment algorithm in computer
  telecommunications}.
\newblock \bibinfo{journal}{Comput. Commun.} \bibinfo{volume}{150},
  \bibinfo{pages}{729--738}.
\newblock \URLprefix \url{https://api.semanticscholar.org/CorpusID:211080814}.
\bibitem[{Xu and Raginsky(2017)}]{Xu2017InformationtheoreticAO}
\bibinfo{author}{Xu, A.}, \bibinfo{author}{Raginsky, M.}, \bibinfo{year}{2017}.
\newblock \bibinfo{title}{Information-theoretic analysis of generalization
  capability of learning algorithms}.
\newblock \bibinfo{journal}{ArXiv} \bibinfo{volume}{abs/1705.07809}.
\newblock \URLprefix \url{https://api.semanticscholar.org/CorpusID:11470350}.
\bibitem[{Xu and Deng(2018)}]{Xu2018DependentEC}
\bibinfo{author}{Xu, H.}, \bibinfo{author}{Deng, Y.}, \bibinfo{year}{2018}.
\newblock \bibinfo{title}{Dependent evidence combination based on shearman
  coefficient and pearson coefficient}.
\newblock \bibinfo{journal}{IEEE Access} \bibinfo{volume}{6},
  \bibinfo{pages}{11634--11640}.
\newblock \URLprefix \url{https://api.semanticscholar.org/CorpusID:3943779}.
\bibitem[{Yamasaki et~al.(2024)Yamasaki, Freire, Seman, Stefenon, Mariani and
  dos Santos~Coelho}]{Yamasaki2024OptimizedHE}
\bibinfo{author}{Yamasaki, M.}, \bibinfo{author}{Freire, R.Z.},
  \bibinfo{author}{Seman, L.O.}, \bibinfo{author}{Stefenon, S.F.},
  \bibinfo{author}{Mariani, V.C.}, \bibinfo{author}{dos Santos~Coelho, L.},
  \bibinfo{year}{2024}.
\newblock \bibinfo{title}{Optimized hybrid ensemble learning approaches applied
  to very short-term load forecasting}.
\newblock \bibinfo{journal}{International Journal of Electrical Power $\&$
  Energy Systems} \URLprefix
  \url{https://api.semanticscholar.org/CorpusID:264535378}.
\bibitem[{Zhang et~al.(2020)Zhang, Hou, Song, Ge and
  Yao}]{Zhang2020RedundancyOH}
\bibinfo{author}{Zhang, C.}, \bibinfo{author}{Hou, Y.}, \bibinfo{author}{Song,
  D.}, \bibinfo{author}{Ge, L.}, \bibinfo{author}{Yao, Y.},
  \bibinfo{year}{2020}.
\newblock \bibinfo{title}{Redundancy of hidden layers in deep learning: An
  information perspective}.
\newblock \bibinfo{journal}{ArXiv} \bibinfo{volume}{abs/2009.09161}.
\newblock \URLprefix \url{https://api.semanticscholar.org/CorpusID:221818661}.
\bibitem[{Zhang et~al.(2023)Zhang, Wu, Zhang, Hu, Fu, Zhou and
  Peng}]{Zhang2023ProvableDF}
\bibinfo{author}{Zhang, Q.}, \bibinfo{author}{Wu, H.}, \bibinfo{author}{Zhang,
  C.}, \bibinfo{author}{Hu, Q.}, \bibinfo{author}{Fu, H.},
  \bibinfo{author}{Zhou, J.T.}, \bibinfo{author}{Peng, X.},
  \bibinfo{year}{2023}.
\newblock \bibinfo{title}{Provable dynamic fusion for low-quality multimodal
  data}, in: \bibinfo{booktitle}{International Conference on Machine Learning}.
\newblock \URLprefix \url{https://api.semanticscholar.org/CorpusID:259075995}.
\bibitem[{Zhang et~al.(2018)Zhang, Zhu, He and Xu}]{Zhang2018ANR}
\bibinfo{author}{Zhang, X.}, \bibinfo{author}{Zhu, Q.}, \bibinfo{author}{He,
  Y.}, \bibinfo{author}{Xu, Y.}, \bibinfo{year}{2018}.
\newblock \bibinfo{title}{A novel robust ensemble model integrated extreme
  learning machine with multi-activation functions for energy modeling and
  analysis: Application to petrochemical industry}.
\newblock \bibinfo{journal}{Energy} \URLprefix
  \url{https://api.semanticscholar.org/CorpusID:117234645}.

\end{thebibliography}

\end{document}